\documentclass[final]{siamltex}
\usepackage{algorithm,algpseudocode}
\usepackage{latexsym, graphicx, epsfig, amsmath, amsfonts,amssymb,bm}
\usepackage{caption, subcaption}
\usepackage{epstopdf}
\usepackage{makecell}
\usepackage{subcaption}
\usepackage{booktabs}
\usepackage{array}
\usepackage{color}
\usepackage{lineno}
\usepackage{url}
\usepackage{diagbox}
\usepackage[normalem]{ulem}
\usepackage{array,multirow,graphicx}

\def\Dt{\Delta t}

\allowdisplaybreaks
\def\be{\begin{equation}}
\def\ee{\end{equation}}

\def\x{\mathbf{x}}

\def\N{\mathbf{N}}

\def\PPh{\boldsymbol{\Phi}}
\def\PPs{\boldsymbol{\Psi}}

\def\R{{\mathbb R}}

\def\I{\mathbf I}

\def\A{\mathbf{A}}

\def\u{\mathbf{u}}
\def\v{\mathbf{v}}
\def\V{\mathbf{V}}
\def\w{\mathbf{w}}

\def\L{\mathcal{L}}

\def\0{\mathbf{0}}

\newcommand{\RevA}{\textcolor{black}}
\newcommand{\RevB}{\textcolor{black}}
\newcommand{\NoRev}{\textcolor{black}}

\title{Principal Component Flow Map Learning of PDEs from incomplete, limited, and noisy data}

\author{Victor Churchill\footnotemark[1]\thanks{Department of Mathematics, Trinity College, Hartford, CT 06119, USA. Email:{\tt victor.churchill@trincoll.edu}. }}

\begin{document}
\maketitle
\begin{abstract}
We present a computational technique for modeling the evolution of dynamical systems in a reduced basis, with a focus on the challenging problem of modeling partially-observed partial differential equations (PDEs) on high-dimensional non-uniform grids. We address limitations of previous work on data-driven flow map learning in the sense that we focus on noisy and limited data to move toward data collection scenarios in real-world applications. %Additionally, we consider three forms of partial data: data are observed only on a subset of the domain, data are observed only on a subset of the state variables, and data are fully observed but a varying parameter is not.
Leveraging recent work on modeling PDEs in modal and nodal spaces, we present a neural network structure that is suitable for PDE modeling with noisy and limited data available only on a subset of the state variables or computational domain. In particular, spatial grid-point measurements are reduced using a learned linear transformation, after which the dynamics are learned in this reduced basis before being transformed back out to the nodal space. This approach yields a drastically reduced parameterization of the neural network compared with previous flow map models for nodal space learning. \RevA{This allows for rapid high-resolution simulations, enabled by smaller training data sets and reduced training times.} %In addition to the mathematical motivation for the neural network structure, we present an extensive set of numerical examples in one-, two-, and three-dimensions to demonstrate the effectiveness of the proposed neural network modeling. %In one example, the method can accurately predict the solution when data are only available in less than half (40\%) of the domain.
\end{abstract}

% main text
\section{Introduction} \label{sec:intro}
Data-driven scientific machine learning has shown promise for modeling time-dependent physical phenomena that do not have a known set of governing equations (e.g. a physics-based model). Such phenomena arise frequently in one-of-a-kind or custom systems, large and small, in applications relevant to our societal well-being and national security as wide ranging as climate modeling, disease preventation, and weapons combustion. Using partial system observations in time and space, state-of-the-art methods train a predictive parameterized model, e.g. a neural network (NN), that can explain unknown dynamics and aid in accelerated simulation for optimization and control, e.g. in digital twin development.

\subsection{Motivation}
When modeling high-dimensional systems, e.g. systems modeled traditionally by partial differential equations (PDEs) with one or more spatial variables observed on a fine spatial grid, NN models require many parameters to account for different system behaviors. In moving from toy problems to real applications, a significant challenge throughout the field is the curse of dimensionality, which is the phenomenon that multiple system observations (training data samples) are required to accurately learn each model parameter without over-fitting. The more complex and high-dimensional a system is, the more parameters are needed in the model, and hence the more system observations are needed for training data. On the other hand, real-world data collection scenarios \RevA{(e.g. storage limitations)} limit the spatial grid on which measurements are made, the time at which observations are made, the state variables that are actually observed, and the amount of data collected. If only a few noisy data are available to train a highly parameterized model, accuracy and generalizability suffer. \RevA{Additionally, high-resolution simulators that generate scientific data often cannot be utilized for analysis locally by different researchers.} Therefore, this paper addresses the immediate need for \RevA{surrogate modeling} methods that require fewer parameters to enable efficient and scalable data-driven modeling of unknown high-dimensional dynamics while maintaining prediction accuracy and stability, especially when partial observations demand memory (i.e. a time history) of inputs. This paper focuses on learning flow maps \RevA{in a linear reduced dimension basis} to reduce the number of model parameters so that limited training data can be used productively \RevA{to enable locally-executable and rapid high-resolution simulations.}

\subsection{Literature Review}

Data-driven learning of unknown partial differential equations (PDEs) has been a prominent area of research for the better part of the last decade now. One approach is governing equation discovery (see e.g. \cite{rudy2019data,rudy2017data,schaeffer2017learning},), whereby a map from the state variables and their spatial derivatives to their time derivatives is constructed. This technique can explicitly reconstruct the governing equations of unknown PDEs from a large dictionary of partial derivative operators using sparsity-promoting regression on solution data collected from the system. In certain circumstances, exact equation recovery is even possible. While the sparsity approach uses data advantageously, it does not apply to equations that are not well-represented by a sparse number of elements in the prescribed dictionary. E.g., a particular fractional derivative may not be in the dictionary, or a term in the PDE may only be approximated by the sum of a large number of terms in the dictionary. %Neural networks (NNs) have also been used to construct this mapping. %See, for example, \cite{long2018pde,long2017pde,lu2021learning,raissi2018deep,raissi2017physics1,raissi2017physics2,sun2019neupde}. 

Another approach which has received significant attention is operator learning. Recent network architecture advances such as DeepONet \cite{lu2021learning}, neural operators \cite{kovachki2023neural,li2020fourier}, and transformer operators \cite{cao2021choose,li2022transformer}, seek to approximate infinite-dimensional operators in a variety of problems related to PDEs such as continuous-time solution operators that map a time $t$ or spatial grid point $x$ to the relevant PDE solution at that point. Because of their general approach, these methods do not exploit the problem of learning the evolution of an autonomous system of PDEs, e.g. to economize the parameterization, and are highly data-intensive. While proven universal approximation theorems suggest economical parameterizations, actual implementations are quite large. Additionally, these networks often have difficulty with accurate long-term prediction accurately when extrapolating time $t$ outside the training set.

This paper focuses on the Flow Map Learning (FML) technique, which constructs an approximate flow map between two discretized system states separated by a short time. If an accurate approximation to the true flow map is constructed, it allows for the definition of a predictive model for the unknown system that marches new initial conditions forward in time. This approach has been shown to be particularly effective when NNs, specifically residual networks (ResNet \cite{he2016deep}), are used to approximate the flow map. Originally designed for ODEs \cite{churchill2023flow,qin2018data}, recent advances were also made with respect to systems traditionally modeled by PDEs, \cite{chen2022deep,churchill2023dnn,WuXiu_modalPDE}. In \cite{WuXiu_modalPDE}, FML of PDEs is conducted in a modal space (i.e. a generalized Fourier basis) by first representing the observations in a reduced basis and then learning the flow map of an associated ODE. In \cite{chen2022deep}, FML of PDEs is conducted in nodal space (i.e. on a spatial grid) using an architecture resembling a forward Euler scheme with finite difference approximations to partial derivatives. Throughout the FML framework, fully-connected networks are used to accommodate complex interactions on non-uniform grids.
%Therefore, especially for partially-observed systems requiring memory (i.e. a time history of the observables),
This induces large parameterizations for high-dimensional state vectors in both modal and nodal FML implementations, requiring even larger training data sets to achieve generalizability and high accuracy over a variety of initial conditions. Perhaps counterintuitively, FML models based on fully-connected NNs penalize high-dimensional observations in this way. The method we propose later is more ambivalent to the observation grid size, and avoids that issue by conducting the learning in a reduced basis. Broadly, there is a need to deal with limited data and non-trivial noise levels.

Of particular importance to this article within the FML framework is \cite{churchill2023dnn}, which extended \cite{chen2022deep} to incomplete or partially-observed PDE systems requiring memory (i.e. a time history). The method there represented a generalization of \cite{chen2022deep} and \cite{FuChangXiu_JMLMC20}, which covered NN modeling of partially-observed ODE systems motivated by the Mori-Zwanzig formulation.\footnote{The same modeling principle was first proposed in \cite{WangRH_2020}, in the context of closure for reduced order modeling using an LSTM structure.}
%Motivated by the celebrated Mori-Zwanzig formulation, \cite{FuChangXiu_JMLMC20} proposed the use of NNs with memory terms to model dynamical systems when only a subset of the state variables are observed. In this paper, we adopt and improve on the simple NN structure from \cite{FuChangXiu_JMLMC20}, which explicitly incorporates memory terms from the measurement data as network inputs, in order to cope with the situation of incomplete data. As a result, the proposed new NN structure is capable of modeling unknown PDE systems using incomplete data.
The main drawback to the memory-based nodal PDE FML framework is its huge parameterization, which requires large amounts of training data to avoid overfitting, and long training and prediction times.

%We address these concerns in two ways:
%\begin{itemize}
%    \item compress spatial measurements to a lower dimension
%    \item approximate an ODE flow map in the reduced dimension
%\end{itemize}

Finally, \RevA{several} other papers, \RevA{\cite{audouze2009reduced,chen2022automated,fresca2021comprehensive,lin2021data,regazzoni2019machine,zeng2024autoencoders}}, have similar goals of \RevA{approximating dynamical system solutions} in a reduced basis, albeit in different contexts. \RevA{In particular, \cite{audouze2009reduced} exclusively focuses on steady-state PDEs. Citations \cite{chen2022automated,zeng2024autoencoders} focus on discovering which, and how many, components from high-dimensional observations are in fact important to system dynamics, while in what follows we focus on working with high-dimensional observations of known states. In \cite{fresca2021comprehensive}, the authors use neural networks to solve a known PDE in a reduced basis.} In \cite{regazzoni2019machine}, authors consider a partially-observed heat equation but observed at only 3 points throughout the domain, with little need for reduction of high-dimensional measurements and focusing more on reducing the dynamics. In \cite{lin2021data}, the authors consider a generalized Koopman framework similarly framed in the context of the Mori-Zwanzig formulation, but only consider a long ($10^7$ time steps), noiseless trajectory from the Lorenz '96 ODE. %In \cite{wang2020recurrent}, the authors focus on integrating LSTM networks into a numerical scheme to close reduced systems, while we focus on learning a reduced order model.

\subsection{Our focus and contributions}

Much like in \cite{churchill2023dnn}, this paper considers data-driven flow map learning of PDEs from incomplete solution data, \RevA{motivated by collection of high-resolution scientific data on either a partial domain or a subset of state variables, e.g. as is done in large-scale climate simulation.} In other words, we consider the case of modeling an unknown PDE system using measurable data that represent partial information of the system. It is important to consider incomplete solution data because in real applications it is often difficult, if not impossible, to collect data of all the state variables of an unknown system in all areas of the relevant domain. It may even be difficult to \emph{identify} all state variables in an unknown system. %Specifically, we consider two cases (as well as their combination): (1) when only a subset of the state variables are observed; and (2) when data of the state variables are available only in a sub-domain of the entire domain. Well if this has been done, why are we writing this paper?

\RevA{Researchers performing analysis of a system on small local machines often do not have direct access to the simulators or systems themselves which are generating this type of high-dimensional scientific data. It is therefore desirable to train a surrogate model from limited data that enables approximate but inexpensive high-resolution simulations that can be executed locally.} Examples in \cite{churchill2023dnn} used up to $10,000$ noise-less training samples to train networks with up to hundreds of thousands of parameters with high accuracy. This demonstrated the not-so-surprising result that with nearly unlimited training time and data generated from densely packed randomized initial conditions, arbitrary accuracy is possible. However, the question now moves to the limited data scenario. As training data comes at a cost in applications, it is desirable to minimize the number of parameters in the model while still capturing accurate system dynamics. In what follows, we \emph{limit data} to no more than 100 samples in all examples, and \emph{add noise}. This combination of factors causes the nodal memory FML models of \cite{churchill2023dnn} to fail, motivating this exploration. 

To reduce the amount of data required to accurately generalize a NN flow map model, the key factor is reducing the size of the model parameterization.\footnote{While there is a growing literature on high over-parameterization to deal with over-fitting when faced with limited data (see e.g. the double descent curve in [1]), the property has not been observed in this context. Indeed, in examples below high over-parameterization still results in over-fitting.}  This depends on both the dimensionality of the observations, i.e. the grid size and the number of observed state variables, as well as the structure of the network itself. The approach of this paper is to:
\begin{itemize}
    \item Reduce the high-dimensional spatial measurements to a few important components by learning common reduced representations for training trajectories;
    \item Reduce the dimensionality of the model (i.e. network parameterization) directly by conducting the dynamics learning in the reduced basis, which requires a lower complexity network architecture due to its ODE form.
\end{itemize}
Therefore, the main contribution of the paper is \emph{a memory-based framework for flow map learning of PDEs in a learned lower-dimensional modal space from incomplete, limited, and noisy data.}
In particular, to reduce the measurements, we introduce unconstrained, constrained, and fixed (or pre-trained) approaches to learning common reduced bases for the training trajectories motivated by principal component analysis (PCA). \RevA{We focus on linear bases to maintain small parametrizations that enable inexpensive simulation, intentionally avoiding data- and training-intensive nonlinear autoencoder-decoders.} To reduce the network parameterization, we derive the existence of an ODE flow map model and additionally propose a matrix form to further reduce the network parameterization. Finally, an adjustment to the overall ResNet structure radically improves the FML strategy's robustness to noise. In a set of \RevA{both linear and nonlinear} numerical examples, we compare these options against each other as well as the existing nodal model from \cite{churchill2023dnn} to demonstrate that our method is indeed highly effective and accurate.
\section{Preliminaries} \label{sec:preliminaries}

%In general, we are interested in constructing effective models for the evolution
%laws behind dynamical data.
%Before the discussion of learning partially-observed PDEs,
In this section, we briefly review the existing NN-based framework for learning the evolution of unknown PDEs in modal space, as well as partially-observed PDEs in nodal space.
Throughout this paper our discussion will be over discrete time instances with a constant time step $\Dt$,
\be \label{tline}
t_0<t_1<\cdots, \qquad
t_{n+1} - t_n = \Dt, \quad \forall n.
\ee
%Generality is not lost with the constant time step assumption. 
We will also use subscript to denote the time variable of
a function, e.g., $\u_n = u(\mathbf{x},t_n)$.

\subsection{FML of PDEs in Modal Space}\label{sec:modal}

The flow map learning approach was first applied to modeling PDEs in \cite{WuXiu_modalPDE}. The idea was that when the solutions of a PDE can be expressed using a fixed basis, the learning can be conducted in modal space. There, data was generated as modal or generalized Fourier space coefficients, such that
\begin{align}\label{eq:projection}
    u(\mathbf{x},t)=\sum_{j=1}^n \hat{u}_j(t)\phi_j(\mathbf{x})%=:u_n(\mathbf{x},t)
\end{align}
where $u_j(t)$ are the coefficients and the $\phi_j(x)$ represent a pre-chosen set of basis functions in a finite-dimensional subspace of the infinite-dimensional space where the solutions live. Letting
\begin{align}
    \hat{\mathbf{u}}(t) = \left(\hat{u}_1(t),\ldots,\hat{u}_n(t)\right)\in\mathbb{R}^{n},
\end{align}
the authors sought to model the flow map of the resulting autonomous ODE system
\begin{align}\label{eq:ODE}
    \frac{d\hat{\mathbf{u}}}{dt}(t) = \mathbf{f}(\hat{\mathbf{u}}(t)),
\end{align}
where $\mathbf{f}:\mathbb{R}^n\rightarrow\mathbb{R}^n$ represents the unknown governing equations for the coefficients.

Per the FML literature on ODEs \cite{churchill2023flow,qin2018data}, the flow map of the system is a function that describes the evolution of the solution. The flow map of \eqref{eq:ODE} depends only
on the time difference but not the actual time, i.e.,
$\hat{\u}_n = \PPh_{t_n-t_s}(\hat{\u}_s)$. Thus, the solution over one time step
satisfies
\be
\hat{\u}_{n+1} = \PPh_{\Dt}(\hat{\u}_n) = \hat{\u}_n + \PPs_{\Dt}(\hat{\u}_n),
\ee
where $\PPs_{\Dt} = \PPh_{\Dt} - \mathbf{I}$, with $\mathbf{I}$ as the
identity operator. When data for the state variables $\hat{\u}$ over the time stencil
\eqref{tline} are available, they can be grouped into sequences separated by
one time step
\be\label{eq:data}
\{\hat{\u}^{(l)}(0), ~~\hat{\u}^{(l)}(\Dt),\ldots,\hat{\u}^{(l)}(K\Delta t)\},\quad l=1,\ldots,L,
\ee
where $L$ is the total number of such data sequences and $K+1$ is the length of each sequence\footnote{The data sequence length $K$ is assumed to be constant for notational convenience. If in fact data are of different lengths, chunks of a minimum sequence length can be taken from longer trajectories.}. These data sequences form the training data set.
Inspired by basic one-step numerical schemes for solving ODEs, one can then model the unknown evolution operator using a residual network (ResNet \cite{he2016deep}) of the form
\be
\mathbf{y}^{out} = \left[\mathbf{I}+\mathbf{N} \right](\mathbf{y}^{in}),
\ee
where $\N:\R^n\to\R^n$ stands for the mapping operator of a standard
feedforward fully connected neural network.
The network is then trained by using the training data set 
and minimizing the recurrent mean squared loss function
\be
\frac1L \sum_{l=1}^L\sum_{k=1}^K \left\| \hat{\u}^{(l)}(k\Dt) - [\I+\N]^k(\hat{\u}^{(l)}(0))\right\|_2^2,
\ee
where $[\mathbf{I}+\mathbf{N}]^k$ indicates composition of the network function $k$ times. This recurrent loss permits the incorporation of longer data lengths and is used to increase the stability of the network approximation over long term prediction.
Once the network is trained to satisfactory accuracy, the trained network thus accomplishes
$$
\hat{\u}^{(l)}(k\Dt) \approx [\mathbf{I}+\N]^k(\hat{\u}^{(l)}(0)), \qquad \forall l=1,\ldots,L, \quad k=1,\ldots,K,
$$
and it can be used as a predictive model
\be \label{ResNet}
\hat{\u}_{n+1} =  \hat{\u}_n + \N(\hat{\u}_n), \qquad n=0,1,2,\dots,
\ee
for initial condition $\hat{\u}(t_0)$. If desired, nodal approximations can be formed via \eqref{eq:projection}.
%This framework was proposed in \cite{qin2018data}, with extensions to parametric systems and time-dependent (non-autonomous) systems (\cite{QinChenJakemanXiu_IJUQ,QinChenJakemanXiu_SISC}).

\subsection{FML of Partially-Observed PDEs in Nodal Space}\label{sec:nodal}

When data of the PDE solutions are available as nodal values over a set of unstructured grid points in physical space, its NN learning is more involved. In \cite{churchill2023dnn}, the NN structure for nodal FML of PDEs from \cite{chen2022deep} was extended to partially-observed systems.

\begin{figure}[htbp]
	\begin{center}
		\includegraphics[width=\textwidth]{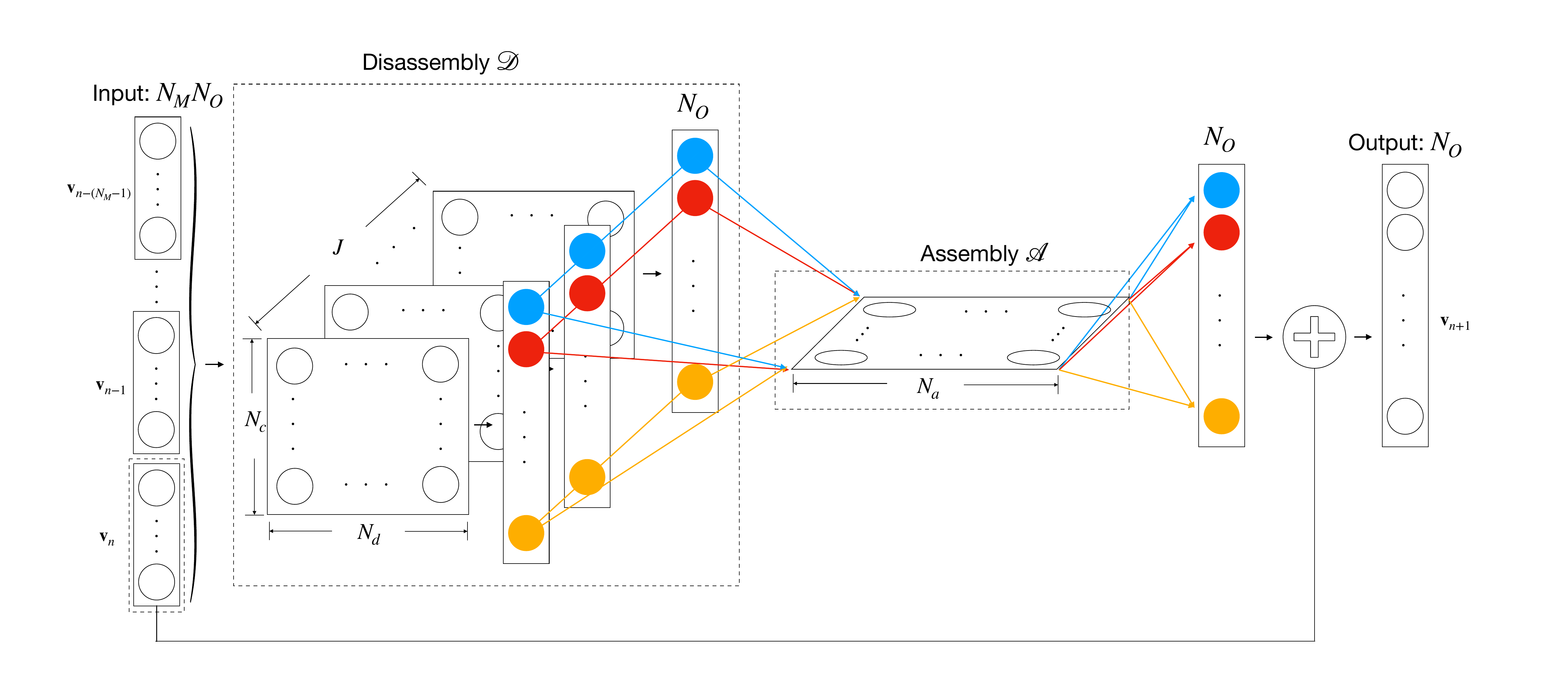}
		\caption{FML model for nodal PDEs with missing information.}
		\label{fig:nodaldiagram}
	\end{center}
\end{figure}

Decomposing the state vector $\u = (\v; \w)$ into observables $\v$ on an $N_O$-dimensional grid and missing information $\w$,\footnote{This general formulation considers missing state variables or missing areas of the domain.} the authors took inspiration from memory-based FML for ODEs \cite{FuChangXiu_JMLMC20}, as well as the Mori-Zwanzig formulation \cite{mori1965,zwanzig1973}, and derived an approximate flow map for $\v$ only using a finite length of time history or memory. Similar to \cite{chen2022deep}, the NN structure of \cite{churchill2023dnn} is based on a simple numerical scheme for solving the PDE, and consists of a set of specialized layers resembling the combination of differential operators involved in the unknown PDE. Specifically, the NN model defines the following mapping,
\be \label{NN_simple}
\v_{n+1} = \v_n + \A(\mathbf{F}_1(\v_n,\ldots,\v_{n-(N_M-1)}), \dots, \mathbf{F}_J(\v_n,\ldots,\v_{n-(N_M-1)})),
\ee
where $\mathbf{F}_1,\dots, \mathbf{F}_J$ are the operators for the disassembly channels and $\A$ the operator for the assembly layer, and $N_M\geq1$ is the number of memory terms in the model. The NN modeling approach was shown to be highly flexible and accurate to learn a variety of PDEs. See \cite{chen2022deep,churchill2023dnn} for more details on this structure and its mathematical properties. Since this configuration can also be viewed as a residual network application, i.e. $\v_{n+1} = [\mathbf{I}+\mathbf{N}](\v_n,\ldots,\v_{n-(N_M-1)})$ where $\mathbf{N} = \mathbf{A}\circ(\mathbf{F}_1,\ldots,\mathbf{F}_J)$, if data sequences of length $N_M+K$
\be\label{eq:data_memory}
\{\v^{(l)}_{n-(N_M-1)},\ldots,\v^{(l)}_{n-1},\v^{(l)}_n,\v^{(l)}_{n+1},\ldots,\v^{(l)}_{n+K}\},\quad l=1,\ldots,L,
\ee
are available, then we can directly apply a recurrent loss scheme
\begin{align}
    \frac1L \sum_{l=1}^L\sum_{k=1}^K\left\|\v^{(l)}_{n+k} - \left[\I + \N\right]^k\left(\v^{(l)}_{n},\ldots,\v^{(l)}_{n-(N_M-1)}\right)\right\|^2,
\end{align}
where $[\mathbf{I}+\mathbf{N}]^k$ indicates composition of the network function $k$ times, using predictions as inputs for $k>1$. The special case of $N_M=1$ corresponds to the standard model from \cite{chen2022deep} for modeling PDEs without missing information (thus no need for memory). As aforementioned, the main issue with this configuration is the large parameterization which in turn is very data thirsty. Figure \ref{fig:nodaldiagram} shows the structure. This technique will serve as the baseline to which we will compare the several proposed models below.

\section{Principal Component Flow Map Learning (PC-FML)} \label{sec:method}

In this main section, we describe a new method for approximating the flow map of partially-observed
%or parameterized
systems of PDEs using NNs. We first set up the problem being considered with appropriate notation, then move on to motivate and discuss the method.

\subsection{Problem Setup and Data Collection}\label{subsec:setup}

Consider an autonomous time-dependent system of PDEs,
\be \label{govern}
\begin{cases}
\mathbf{u}_t = \L(\mathbf{u}), \quad &(x,t) \in \Omega \times \mathbb R^+,
\\
{\mathcal B} (\mathbf{u}) = 0, \quad &(x,t) \in \partial \Omega \times \mathbb
R^+,\\
\mathbf{u}(x,0) = \mathbf{u}_0(x), \quad & x \in \bar{\Omega}, 
\end{cases}
\ee
where $\mathbf{u}$, the vector of state variables, has $N_{var}$ elements, $\Omega\subset\mathbb{R}^{N_d}$, $N_d=1,2, 3$, is the physical domain, and $\mathcal{L}$ and $\mathcal{B}$ stand for the PDE operator and boundary condition operator, respectively, of the entire system. We assume that the PDE is unknown.

We assume that solution data of the PDE are only partially observable. First, let $\mathbf{u} = (\mathbf{v};\mathbf{w})$, where $\mathbf{v}$ contains the $N_{obs}$ state variables with available data and $\mathbf{w}$ contains the $N_{var}-N_{obs}$ unobserved subset of state variables. Furthermore, we assume that $\mathbf{v}$ is observed at a set of points,
\be\label{eq:grid}
X_{N_{grid}} = \{x_1,\ldots,x_{N_{grid}}\}\subset \Omega.
\ee
The set $X_{N_{grid}}$ is quite a general grid that may in fact be a discretization of some proper subset of $\Omega$ such that $\text{Conv}(X_{N_{grid}}) \subsetneq \Omega$ where $\text{Conv}(X_{N_{grid}})$ is the convex hull of $X_{N_{grid}}$. We emphasize that the difference can be nontrivial and also represent an incomplete or partially-observed measurement, i.e. $\left|\Omega\setminus\text{Conv}(X_{N_{grid}})\right| = \mathcal{O}(1)$ where $|\cdot|$ represents the Lebesgue measure.

Using vector notation, we concatenate 
\be\label{eq:concat}
\mathbf{V}(t) = (V_1(t),\ldots,V_{N_{obs}}(t))^T\in\mathbb{R}^{N_{obs}N_{grid}},
\ee
where each component of the subset of state variables is given by
\be
V_j(t) = (v_j(x_1,t),\ldots,v_j(x_{N_{grid}},t))\in\mathbb{R}^{N_{grid}}, \quad j=1,\ldots,N_{obs}.
\ee
%As in \cite{chen2022deep}, we note that the DNN structure that follows will allow us to simply concatenate the state variable solutions on grids (or arbitrarily permute these grids if data is collected in a different manner).
These state variables are observed at discrete time instances \eqref{tline} such that the set
\begin{align}\label{eq:data_pdememory}
\left\{\mathbf{V}^{(l)}_{n-(N_{mem}-1)},\ldots,\mathbf{V}^{(l)}_n,\mathbf{V}^{(l)}_{n+1},\ldots,\mathbf{V}^{(l)}_{n+N_{rec}}\right\},\quad l=1,\ldots,N_{traj},
\end{align}
represents all available data similar to \eqref{eq:data_memory} where $N_{mem}\ge1$ is the memory length, $N_{rec}$ is the recurrent loss parameter, and $N_{traj}$ is the total number of trajectories available. Note that we continue to use the compact notation, $\mathbf{V}_n = \mathbf{V}(t_n)$. Accounting for each of the $N_{traj}$ training samples with length of each vector $N_{obs} N_{grid}$ observed over the memory and recurrent length $N_{mem}+N_{rec}$, the dataset is of size $N_{traj}\times (N_{obs} N_{grid}) \times (N_{mem}+N_{rec})$. From the data \eqref{eq:data_pdememory}, our goal is to model the flow map
\begin{align}\label{eq:nodalFML}
    \mathbf{V}(\x,t+\Delta t)=\Phi_{\Delta t}(\mathbf{V}(\x,t))
\end{align}
for the reduced system of observed variables $\mathbf{V}$ on the grid $\x$. This construction is fundamentally different than \cite{WuXiu_modalPDE}, which sought to model the modal coefficient flow map \eqref{ResNet} whereas this study models the nodal flow map \eqref{eq:nodalFML}.

Additionally, we highlight that the data collection scenario is different than in \cite{WuXiu_modalPDE}. There, nodal data were generated from modal space coefficients by first choosing a basis $\{\phi_j(\x)\}_{j=1}^n$ as in \eqref{eq:projection}, implying a known invertible map between modal and nodal measurements. Here the spatial measurement data are directly collected from physical or nodal space with no knowledge of ideal modal spaces, with finding an appropriate basis one of the premises of this exploration. Neither the existence of nor the form of particular basis functions of a map between nodal and modal measurements are known and therefore must be discovered.

%Therefore, we are concerned with approximating $\varepsilon_\Delta:\mathbb{R}^{NN_{obs}}\rightarrow\mathbb{R}^{NN_{obs}}$, which is $\varepsilon_\Delta \left[\mathbf{V}(\mathbf{x},t)\right] = \mathbf{V}(\mathbf{x},t+\Delta t)$. As demonstrated in \cite{churchill2023dnn}, because $\mathbf{V}$ represents a partial observation of the entire system, in fact this evolution operator depends on a time history of $\mathbf{V}$. That is, $\varepsilon\left[\mathbf{V}_n,\ldots,\mathbf{V}_{n-(N_{mem}-1)}\right]$ where again $N_{mem}$ is an integer prescribing sufficient time history.

\subsection{Mathematical Motivation}\label{subsec:theory}

\subsubsection{Reduced basis for nodal measurements}
We begin reducing the parameter space of the NN model by reducing the dimension of the observations. For ease of presentation, we define $N_{full} = N_{obs}N_{grid}$ and consider the observable state vector $\V_n\in\mathbb{R}^{N_{full}}$ of an unknown PDE at time $t_n$ such as in \eqref{govern}, recalling that $N_{mem}$ such vectors represent the entire input.

We assume that there exists a set of basis functions in the physical domain $\Omega$,
\begin{align}\label{eq:basis}
\{b_{j}(\mathbf{x})\}_{j=1}^{N_{full}},
\end{align}
such that solutions $\V_n$ admit an accurate $N_{red}$-term series approximation
\begin{align}\label{eq:approx}
    \V_n\approx\tilde{\mathbf{V}}_n &= %\left(\sum_{j=1}^{N_{red}} \hat{v}_{1,j}(t)b_{1,j}(\mathbf{x}),\ldots,\sum_{j=1}^{N_{red}} \hat{v}_{N_{obs},j}(t)b_{N_{obs},j}(\mathbf{x})\right)^T
    % \tilde{\mathbf{v}}_n &= \sum_{j=1}^{N_b} \hat{v}_j(t_n)b_j(\mathbf{x})
%\end{align}
%\begin{align}\label{eq:vhatbasis}
 %\V(\x,t) \approx
 \sum_{j=1}^{N_{red}} \hat{V}_j(t)b_j(\x),
\end{align}
with $N_{red}<N_{full}$, incurring error
\begin{align}\label{eq:representation_error}
\delta_n^{(l)}=\left\|\mathbf{V}_n^{(l)}-\tilde{\mathbf{V}}^{(l)}_n\right\|^2_2,
\end{align}
for $l=1,\ldots,N_{traj}$ and $n=1,\ldots,N_{mem}+N_{rec}$ as in \eqref{eq:data_pdememory}. Generally, if $N_{red}$ were fixed, we might seek an optimal basis \RevB{that accurately approximates training data by minimizing over $\delta_n^{(l)}$ for all trajectories and time steps in the training data.} We note that an optimal basis may be significantly different from those generating each training instance's exact finite-term series representation of the finite-dimensional grid measurements.

More simply put, we assume that the solutions to the unknown PDE admit a common linear approximation. As will be demonstrated in the numerical examples, this need not be a particularly accurate approximation and at times (especially with noisy measurements) higher values for $N_{red}$ are not advantageous. This assumption limits the study to functions well-approximated by a common linear basis. Of course, there are solutions to PDEs that do not admit accurate linear reduced order approximations, e.g. traveling waves with discontinuous initial conditions or linear transport \cite{greif2019decay}. The viability of solutions to PDEs having accurate linear reduced order approximations can be quantified by the decay of the Kolmogorov n-width \cite{pinkus2012n}. We plan to consider more general bases in future work.

%We stress that this is a mild assumption.

%We note that even if we consider infinite-dimensional functions as in \cite{WuXiu_modalPDE}, this is a mild assumption. For example, a Fourier series is adequate.\footnote{E.g., for bi-variate function $f(x,y)$ defined on $\Omega = [-\pi,\pi]\times[-\pi,\pi]$
%\begin{align}
%f(x,y) &= \sum_{j,k\in\mathbb{Z}} c_{j,k}e^{ijx} e^{iky} \text{ with }
%c_{j,k} = \frac{1}{4\pi^2}\int_{-\pi}^\pi \int_{-\pi}^\pi f(x,y) e^{-ijx}e^{-iky}dxdy.
%\end{align} Alternatively, if $f$ is an $\alpha$-H\"older continuous function, its Fourier series is uniformly convergent in $L^2$ norm.}

Similar to the definition of $\mathbf{V}$ in \eqref{eq:concat}, let the concatenation of the modal expansion coefficients be
\begin{align}
    \hat{\mathbf{V}}_n = \left(\hat{V}_{1}(t_n),\ldots,\hat{V}_{N_{red}}(t_n)\right)^T.
\end{align}
Considering these definitions, we define two linear transformations
\begin{align}\label{eq:pi}
\begin{split}
\Pi_{in}&\in\mathbb{R}^{N_{red}\times N_{full}}, \text{ such that } \Pi_{in}\V_n = \hat{\V}_n, \\
\\
\Pi_{out}&\in\mathbb{R}^{N_{full}\times N_{red}}, \text{ such that } \Pi_{out}\hat{\V}_n = \tilde{\V}_n,
\end{split}
\end{align}
which transform the nodal measurements to the reduced modal space and back out.

A focus of the ensuing technique is on finding matrices $\Pi_{in}$ and $\Pi_{out}$, that balance minimization of the error $\delta_n^{(l)}$ and the reduced dimension $N_{red}$ over the entire training dataset \eqref{eq:data_pdememory}. As described in Section \ref{subsec:training}, several approaches are taken, motivated by principal component analysis (PCA), also known as principal orthogonal decomposition (POD). E.g., perhaps the most straightforward approach is to use fixed transformations based on the PCA weight matrix which can be pre-computed from the training data. This would enforce an accurate approximation of $\V_n$ by $\Pi_{out}\Pi_{in}\V_n$ along with $\Pi_{in}=\Pi_{out}^T$ and $\Pi_{out}$ having orthogonal columns.

%\begin{align}
%v_i(\mathbf{x},t) &= \sum_{j=1}^\infty \hat{v}_{i,j}(t)b_{i,j}(\mathbf{x})\\
%\mathbf{v}_n &= \sum_{j=1}^\infty \hat{v}_{j}(t_n)b_j(\mathbf{x})\\
%\mathbf{V}_n &= \left(\sum_{j=1}^{N_{grid}} \hat{v}_{1,j}(t_n)b_{1,j}(\mathbf{x}),\ldots,\sum_{j=1}^{N_{grid}} \hat{v}_{N_{obs},j}(t_n)b_{N_{obs},j}(\mathbf{x})\right)^T
%\end{align}

\subsubsection{Partial reduced dynamics}

%Consider a system of ODEs,
%\begin{align}
%\frac{d\hat{\mathbf{u}}}{dt} = \mathbf{f}(\hat{\mathbf{u}}),\quad \hat{\mathbf{u}}(0) = \hat{\mathbf{u}}_0,
%\end{align}
%where $\hat{\mathbf{u}}\in\mathbb{R}^{nN_b}$ are the state variables. We assume that the form of the governing equations, which manifests itself via $\mathbf{f}:\mathbb{R}^{nN_b}\rightarrow\mathbb{R}^{nN_b}$, is unknown. Let $\hat{\mathbf{u}} = (\hat{\mathbf{v}};\hat{\mathbf{w}})$ where $\hat{\mathbf{v}}\in\mathbb{R}^{mN_b}$ is the subset of the state variables with available data, and $\hat{\mathbf{w}}\in\mathbb{R}^{(n-m)N_b}$ is the unobserved subset of the state variables.

%Additionally, let $\mathbf{u} = (\mathbf{v};\mathbf{w})$, where $\mathbf{v}$ contains the $m$ state variables with available data and $\mathbf{w}$ contains the $n-m$ unobserved subset of state variables. Our goal is to model the effective governing equations for the reduced system of observed variables $\mathbf{v}$. When $\mathbf{u}$ is not fully observed, the corresponding ODE system $\frac{d\hat{\mathbf{u}}}{dt} = \mathbf{f}(\hat{\mathbf{u}})$ is partially observed.

Motivated by the modal approximation \eqref{eq:approx},  there exists an unknown formal system of ODEs for the modal expansion coefficients $\hat{\V}\in\mathbb{R}^{N_{red}}$ of the form,
\begin{align}\label{eq:vhatode}
\frac{d\hat{\V}}{dt} = \mathbf{f}(\hat{\V}(t)),
\end{align}
the solution of which has an approximate correspondence to the solution of the PDE in $\mathbf{V}$. This permits learning the evolution of the system in $\mathbb{R}^{N_{red}}$ instead of $\mathbb{R}^{N_{full}}$.
When the governing PDE is unknown, the system for $\hat{\mathbf{V}}$ is unknown as well. Since $\mathbf{V}$ represent partial observations of the entire PDE system, $\hat{\mathbf{V}}$ represent partial observations of some entire system, say for $\hat{\mathbf{U}}$, the modal coefficients of some approximation of the full system.

We seek to discover the dynamics for only the observables $\hat{\mathbf{V}}$ in this reduced space, the evolution of which is non-autonomous per the Mori-Zwanzig formalism \cite{mori1965transport,zwanzig1973nonlinear}, and follows a generalized Langevin equation.
%\begin{align}
%\frac{d}{dt}\hat{\mathbf{V}}(t) = \mathbf{R}(\hat{\mathbf{V}}(t))+\int_0^t \mathbf{K}(\hat{\mathbf{V}}(t-s),s)ds+\mathbf{F}(t,\hat{\mathbf{U}}_0).
%\end{align}
%The $\mathbf{R}$ term depends only on the reduced variables $\hat{\mathbf{v}}$ at the current time and is Markovian. The second term, known as the memory, depends on the reduced variables $\hat{\mathbf{v}}$ at all times, from the initial time $s=0$ to current time $s=t$. Its integrand involves the memory kernel $\mathbf{K}$. The last term is called orthogonal dynamics, which depends on the entire initial state $\hat{\mathbf{u}}_0$, is treated as noise.
Referring to \cite{FuChangXiu_JMLMC20} for details, a time-discretized finite memory approximation models the coefficient flow map as
\begin{align}\label{eq:odememory}
    \hat{\mathbf{V}}_{n+1} = \hat{\mathbf{V}}_n+\mathcal{M}\left(\hat{\mathbf{V}}_n,\ldots,\hat{\mathbf{V}}_{n-(N_{mem}-1)}\right).
\end{align}
where $N_{mem}\ge1$ is the number of memory terms in the model\footnote{The number of memory steps $N_{mem}$ is problem-dependent on both the length of the time-step and the dynamics. We plan to evaluate its a priori optimization in a future investigation.}, and $\mathcal{M}:\mathbb{R}^{N_{mem}N_{red}}\rightarrow\mathbb{R}^{N_{red}}$ is a fully-connected NN.

Since data is collected in the nodal space, all of the modal \emph{input} terms in \eqref{eq:odememory} can be viewed as nodal terms by factoring out $\Pi_{in}$ per \eqref{eq:pi}, retrieving
\begin{align}
    \Pi_{in}\mathbf{V}_{n+1} = \Pi_{in}\mathbf{V}_n+\mathcal{M}\left(\Pi_{in}\mathbf{V}_n,\ldots,\Pi_{in}\mathbf{V}_{n-(N_{mem}-1)}\right).
\end{align}
Furthermore, we can transform the model \emph{outputs} to $\mathbb{R}^{N_{full}}$ by applying $\Pi_{out}$,
\begin{align}
    \Pi_{out}\Pi_{in}\mathbf{V}_{n+1} = \Pi_{out}\Pi_{in}\mathbf{V}_n+\Pi_{out}\mathcal{M}\left(\Pi_{in}\mathbf{V}_n,\ldots,\Pi_{in}\mathbf{V}_{n-(N_{mem}-1)}\right).
\end{align}
This suggests the following model for the nodal space flow map as
\begin{align}\label{eq:model}
    \bar{\mathbf{V}}_{n+1} = \Pi_{out}\Pi_{in}\bar{\mathbf{V}}_{n}+\Pi_{out}\mathcal{M}\left(\Pi_{in}\bar{\mathbf{V}}_{n},\ldots,\Pi_{in}\bar{\mathbf{V}}_{n-(N_{mem}-1)}\right),
\end{align}
where $\mathcal{M}$ approximates the modal space coefficient flow map, and the bar notation \RevB{$\bar{\mathbf{V}}$ denotes that model inputs and outputs are approximations of $\mathbf{V}$.}
%We provide the following error analysis, not with the goal of actually quantifying the error in our examples, but in order to analyze the types of error incurred and their intuitive explanations.
%\begin{proposition}\label{prop}
    %Given the approximation method \eqref{eq:model}, the error at time $t_n$ is given by
    %\begin{align}
     %   e(t_n) &= \varepsilon^{proj}_{n-1}+\left\|\Pi_{out}\Pi_{in}\right\|_2e(t_{n-1})+\varepsilon^{\Delta}_{n} + \left\|\Pi_{out}\right\|_2\left\|\hat{\varepsilon}-\mathcal{M}\right\|_2\left\|\Pi_{in}\right\|_2\sum_{k=1}^{N_{mem}}e(t_{n-k})
    %\end{align}
%\end{proposition}
%The proof is in the appendix. Corresponding to the four terms in \ref{prop}, the approximation error can be categorized into four groups:
%\begin{enumerate}
 %   \item Error incurred by projecting $\mathbf{V}_{n-1}$ into and out of the reduced space
  %  \item Local truncation error at $t_{n-1}$ scaled by size of $\Pi_{out}\Pi_{in}$
   % \item Error incurred in reducing dynamics to coefficient space
    %\item Error incurred by memory terms with approximate coefficient flow map
%\end{enumerate}
%It is possible to eliminate the first error and the factor on the second by using just $\bar{\mathbf{V}}_{n-1}$ as the starting point to which to add the residual. However,
After extensive numerical experimentation, we find that the transformation of $\bar{\V}_n$ into and out of the reduced space is essential to dealing with noisy nodal measurements.
Essentially, $\Pi_{out}\Pi_{in}$ functions as a denoiser. If measurements are noise-free, it can be removed.

\begin{figure}[htbp]
	\begin{center}
		\includegraphics[width=\textwidth]{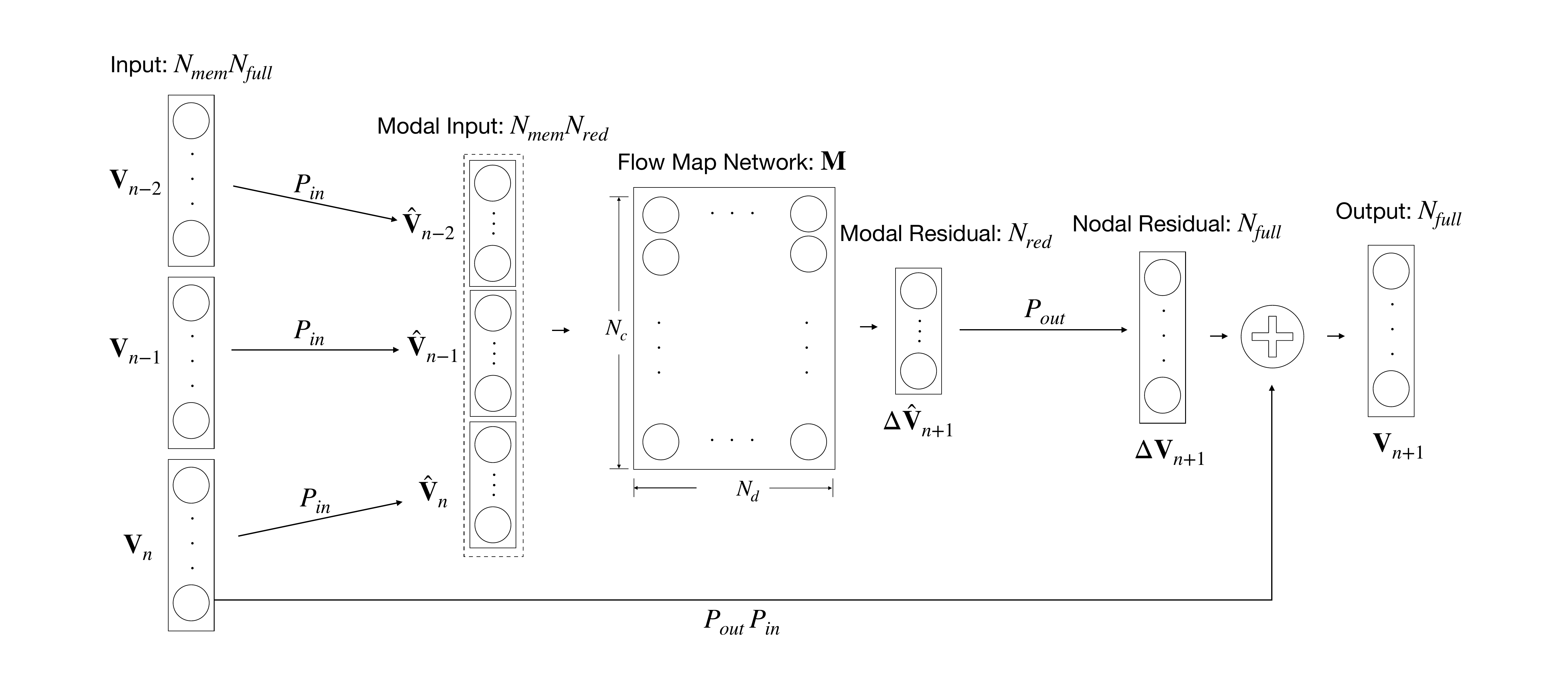}
		\caption{The proposed PC-FML network structure, with $N_{mem}=3$.}
		\label{fig:reduced_diagram}
	\end{center}
\end{figure}

\subsection{Numerical Approach}\label{subsec:newstructure}
Per the motivation above, we propose to model the approximate flow map \eqref{eq:model} using NNs. That is, network operations will learn the operators $\Pi_{in}$, $\Pi_{out}$ and $\mathcal{M}$. The overall network operation, shown in Figure \ref{fig:reduced_diagram}, consists of:
\begin{itemize}
\item \underline{Input reduction:} each of the $N_{full}$-length state vectors in the $N_{mem}$-length memory period are reduced to modal space via a matrix $P_{in}\in\mathbb{R}^{N_{red}\times N_{full}}$, after which they are concatenated into a $N_{mem}N_{red}$-length vector.
\item \underline{Flow map network:} a fully-connected neural network $\mathbf{M}:\mathbb{R}^{N_{mem}N_{red}}\rightarrow\mathbb{R}^{N_{red}}$ maps the reduced time history to the modal residual.
\item \underline{Residual expansion:} expands the modal residual to a nodal residual via a matrix $P_{out}\in\mathbb{R}^{N_{full}\times N_{red}}$.
\item \underline{Skip connection:} adds the current nodal time step having been projected into the reduced space.
\end{itemize}
Together, the network operation is:
\begin{align}\label{eq:networkmodel}
\bar{\V}_{n+1} &= P_{out}P_{in}\bar{\V}_n+P_{out}\mathbf{M}\left(P_{in}\bar{\V}_n,P_{in}\bar{\V}_{n-1},\ldots,P_{in}\bar{\V}_{n-(N_{mem}-1)}\right)
\end{align}
We will refer to this as Principal Component Flow Map Learning (PC-FML) model. Generally, the components $P_{in}$, $P_{out}$, and $\mathbf{M}$ represent the trainable parameters in the network. However, Section \ref{subsec:training} provides several options for their specification. We note that compared with the partially-observed PDE flow map learning of \cite{churchill2023dnn}, modeling via \eqref{eq:networkmodel} has the potential to significantly reduce the number of parameters in the model. E.g., $P_{in}$ and $P_{out}$ represent $2N_{full}N_{red}$ parameters as they are matrices, while at minimum $\mathbf{M}$ represents $N_{mem}N_{red}^2$ parameters. Meanwhile, in \cite{churchill2023dnn}, networks had at least a multiple (typically $\ge5$ times) of $N_{mem}N_{full}^2$ parameters.

%Note: $P_{out}$ is intentionally applied separately to the skip connection and residual. After extensive numerical testing, this minimized the chance of cancellation between the two quantities as opposed to adding and then applying $P_{out}$ just once.

\subsection{Determining reduced order dimension}\label{sec:chooserom}

Given the $N_{traj}\times N_{full}\times (N_{mem}+N_{rec})$ data tensor in \eqref{eq:data_pdememory}, the PC-FML approach requires specification of the modal dimension $N_{red}$. Assemble and reshape the data into an $N_{traj}(N_{mem}+N_{rec})\times N_{full}$ matrix, say $D$. This matrix represents in rows the total number of observations in the training dataset and in columns the dimension of the observations. Compute its singular value decomposition (SVD): $D=U\Sigma V^T$. 
%\begin{align}
    %D&={U}{\Sigma}V^T
%\end{align}
%where
%\begin{itemize}
 %   \item $U\in\mathbb{R}^{N_{data}(N_{mem}+N_{rec})\times N_{data}(N_{mem}+N_{rec})$ is orthogonal,
  %  \item $\Sigma\in\mathbb{R}^{N_{data}(N_{mem}+N_{rec})\times N_{grid}}$ is non-negative diagonal (decaying),
   % \item and $V\in\mathbb{R}^{N_{full}\times N_{full}}$ has orthonormal columns.
%\end{itemize}
Looking at the decay of the singular values, we can determine an appropriate reduced order dimension $N_{red}$ (or if one exists). By comparing $D$ and $D_{red}=U_{red}\Sigma_{red}V_{red}^T$, the associated rank-$N_{red}$ truncated SVD approximation, we can further examine if $N_{red}$ is appropriate. E.g., we may check the largest absolute deviation of the approximation over all trajectories. Different values of $N_{red}$ can be tested for accuracy as a small pre-computational cost which has no requirements on the amount of training data.

%\begin{figure}[htbp]
%	\begin{center}
%		\includegraphics[width=.32\textwidth]{./Figures/ex1_dimension}
%        \includegraphics[width=.32\textwidth]{./Figures/burgers_dimension}
%        \includegraphics[width=.32\textwidth]{./Figures/ex2_dimension}
%        \includegraphics[width=.32\textwidth]{./Figures/swe_dimension}
%        \includegraphics[width=.32\textwidth]{./Figures/ex3_dimension}
%        \includegraphics[width=.32\textwidth]{./Figures/ns_dimension}
%		\caption{Determining reduced order dimension.}
%		\label{fig:dimension}
%	\end{center}
%\end{figure}

\begin{figure}[htbp]
    \centering
    % Example 1: Heat
    \begin{subfigure}[b]{0.31\textwidth}
        \centering
        \includegraphics[width=\textwidth]{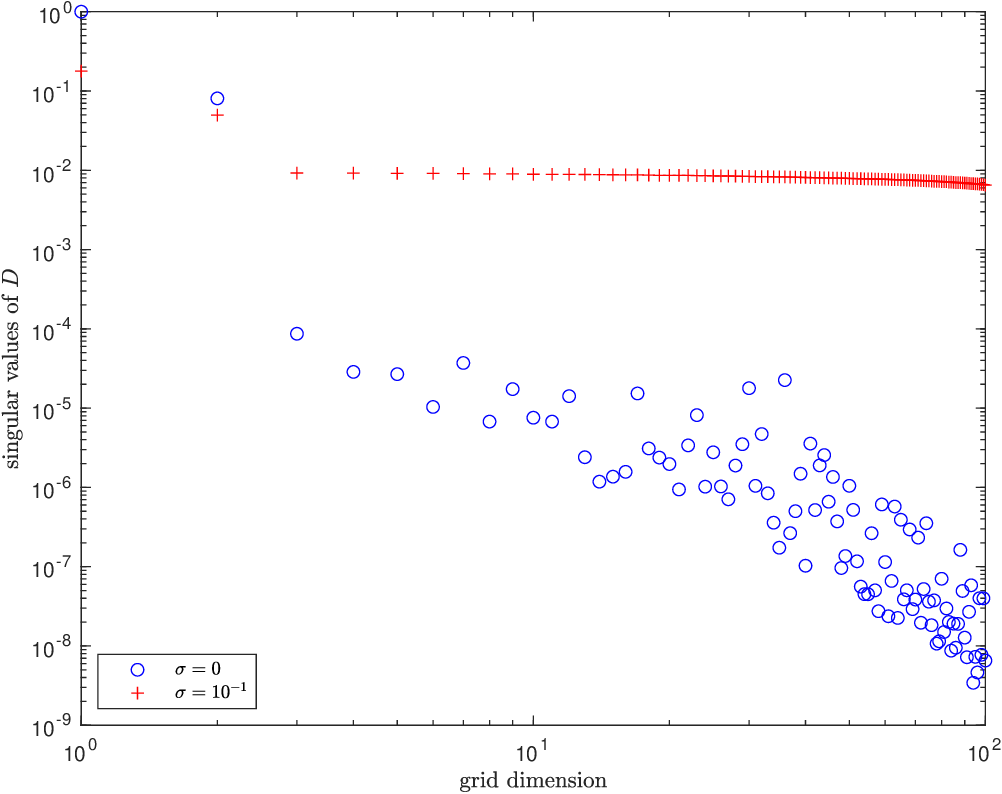}
        \caption{Ex. 1: Heat}
        % \label{fig:heat}
    \end{subfigure}
    % Example 2: Burgers'
    \begin{subfigure}[b]{0.31\textwidth}
        \centering
        \includegraphics[width=\textwidth]{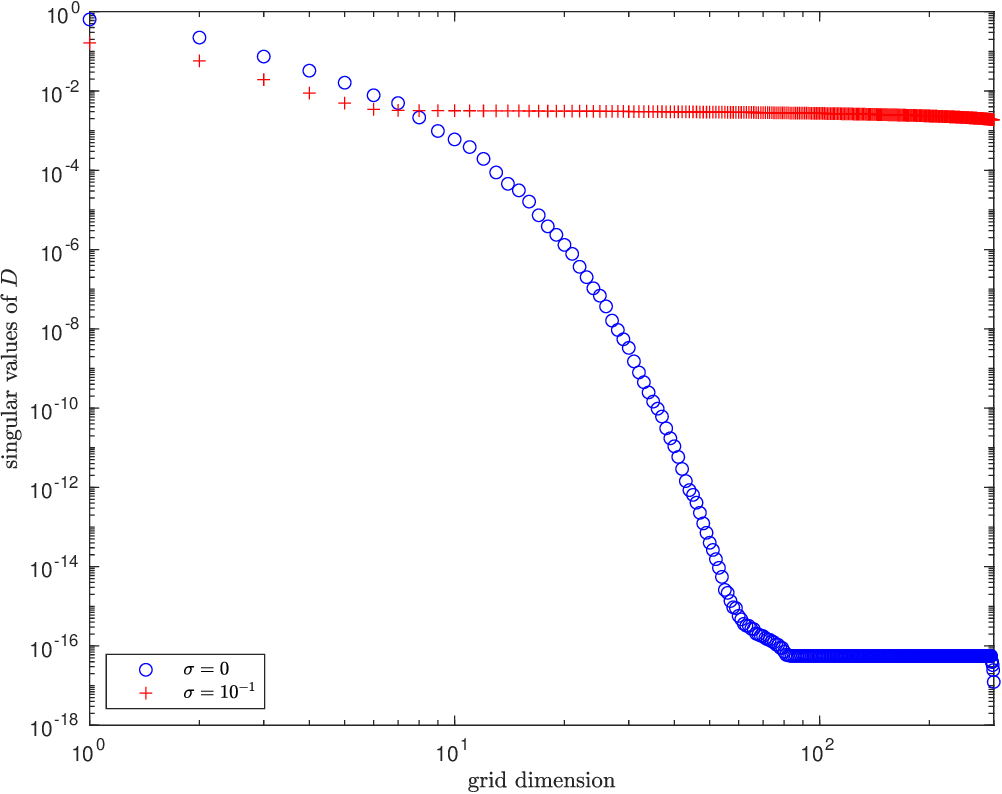}
        \caption{Ex. 2: Burgers'}
        % \label{fig:burgers}
    \end{subfigure}
    % Example 3: Wave 1D
    \begin{subfigure}[b]{0.31\textwidth}
        \centering
        \includegraphics[width=\textwidth]{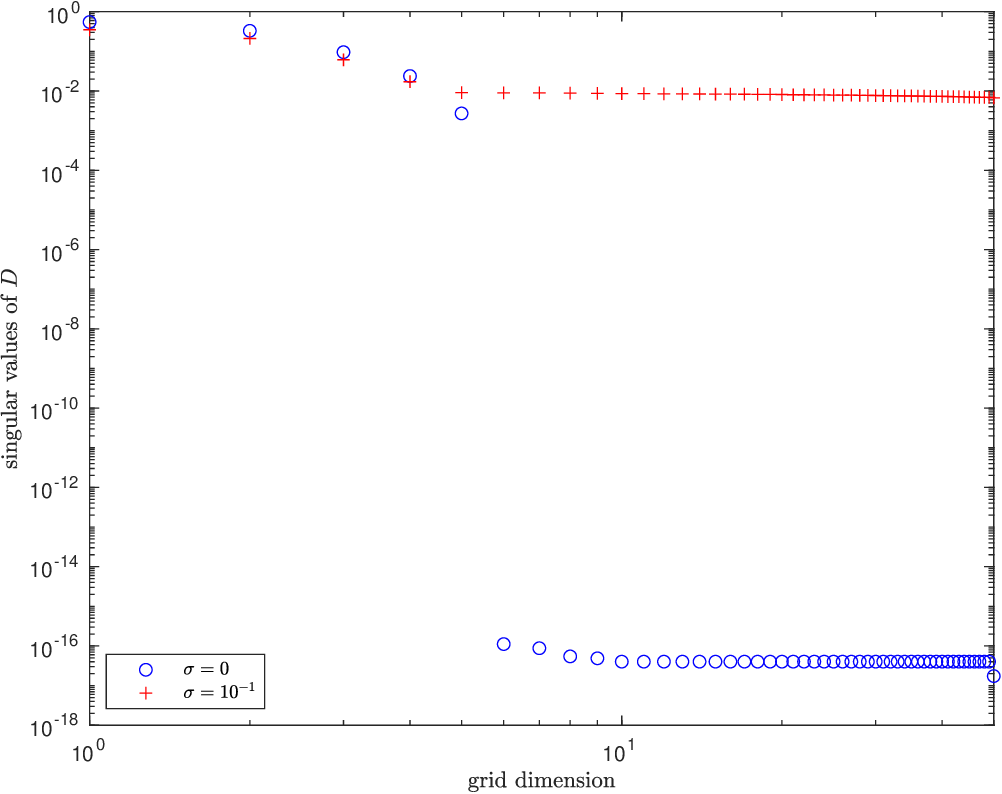}
        \caption{Ex. 3: Wave 1D}
        % \label{fig:wave1d}
    \end{subfigure}
    % Example 4: Shallow Water
    \begin{subfigure}[b]{0.31\textwidth}
        \centering
        \includegraphics[width=\textwidth]{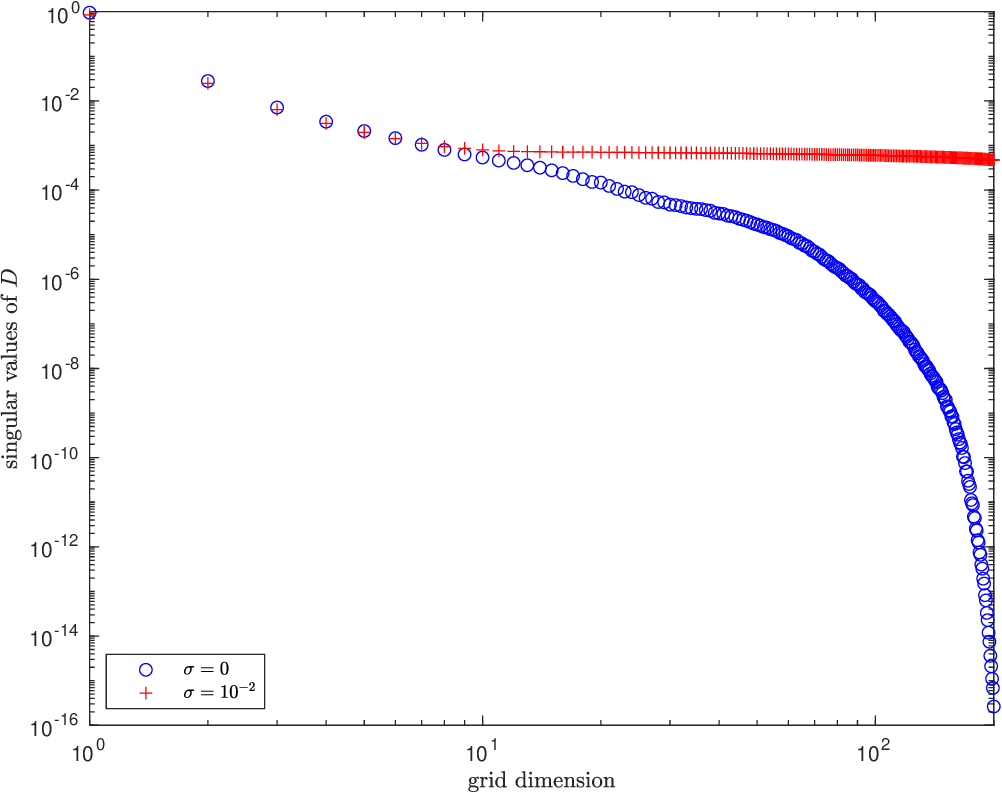}
        \caption{Ex. 4: Shallow Water}
        % \label{fig:shallow_water}
    \end{subfigure}
    % Example 5: Wave 2D
    \begin{subfigure}[b]{0.31\textwidth}
        \centering
        \includegraphics[width=\textwidth]{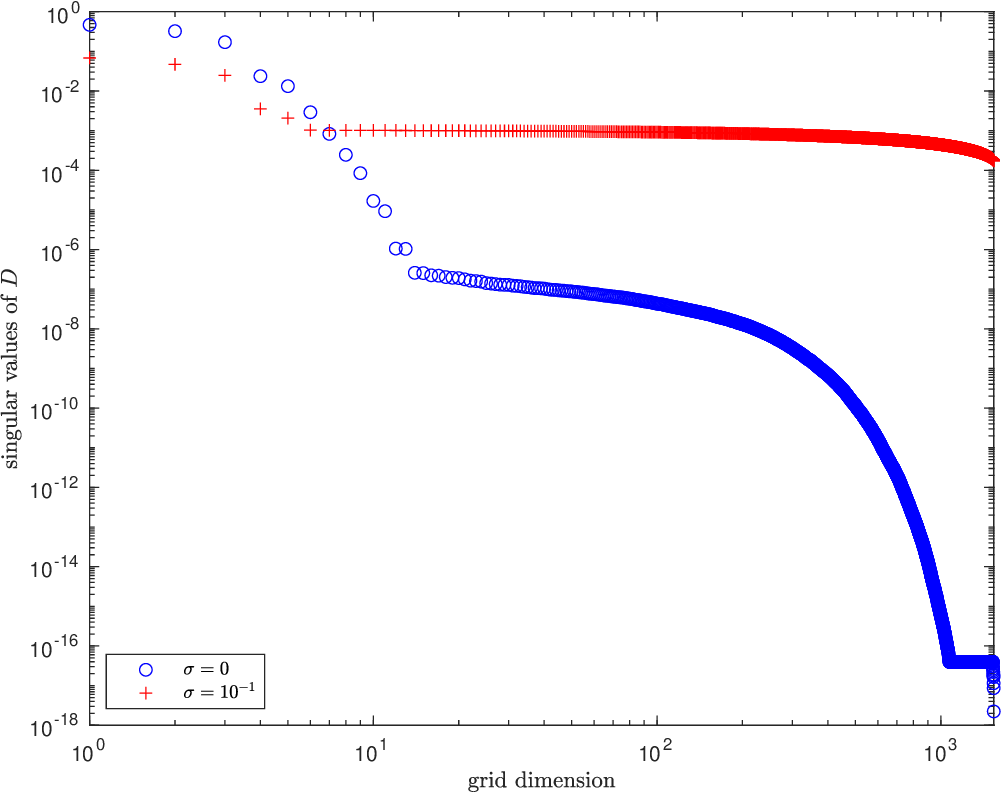}
        \caption{Ex. 5: Wave 2D}
        % \label{fig:wave2d}
    \end{subfigure}
    % Example 6: Navier-Stokes
    \begin{subfigure}[b]{0.31\textwidth}
        \centering
        \includegraphics[width=\textwidth]{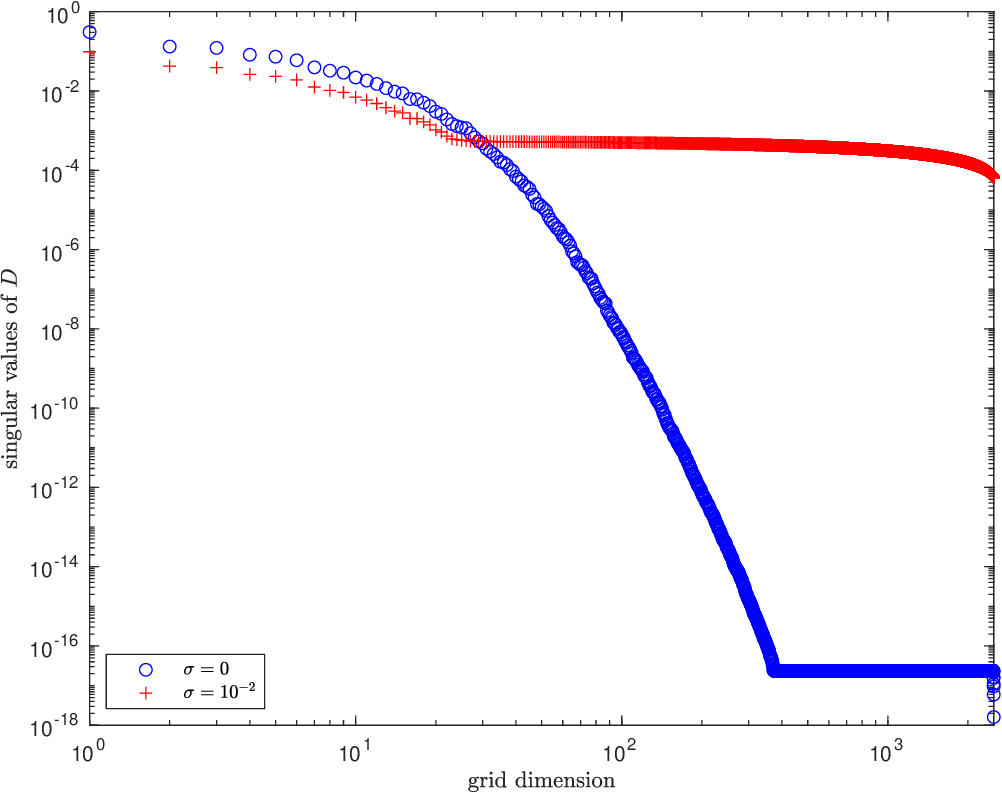}
        \caption{Ex. 6: Navier-Stokes}
        % \label{fig:navier_stokes}
    \end{subfigure}
    \caption{Determining reduced order dimension.}
    \label{fig:dimension}
\end{figure}

\begin{figure}[htbp]
	\begin{center}
		\includegraphics[width=.99\textwidth]{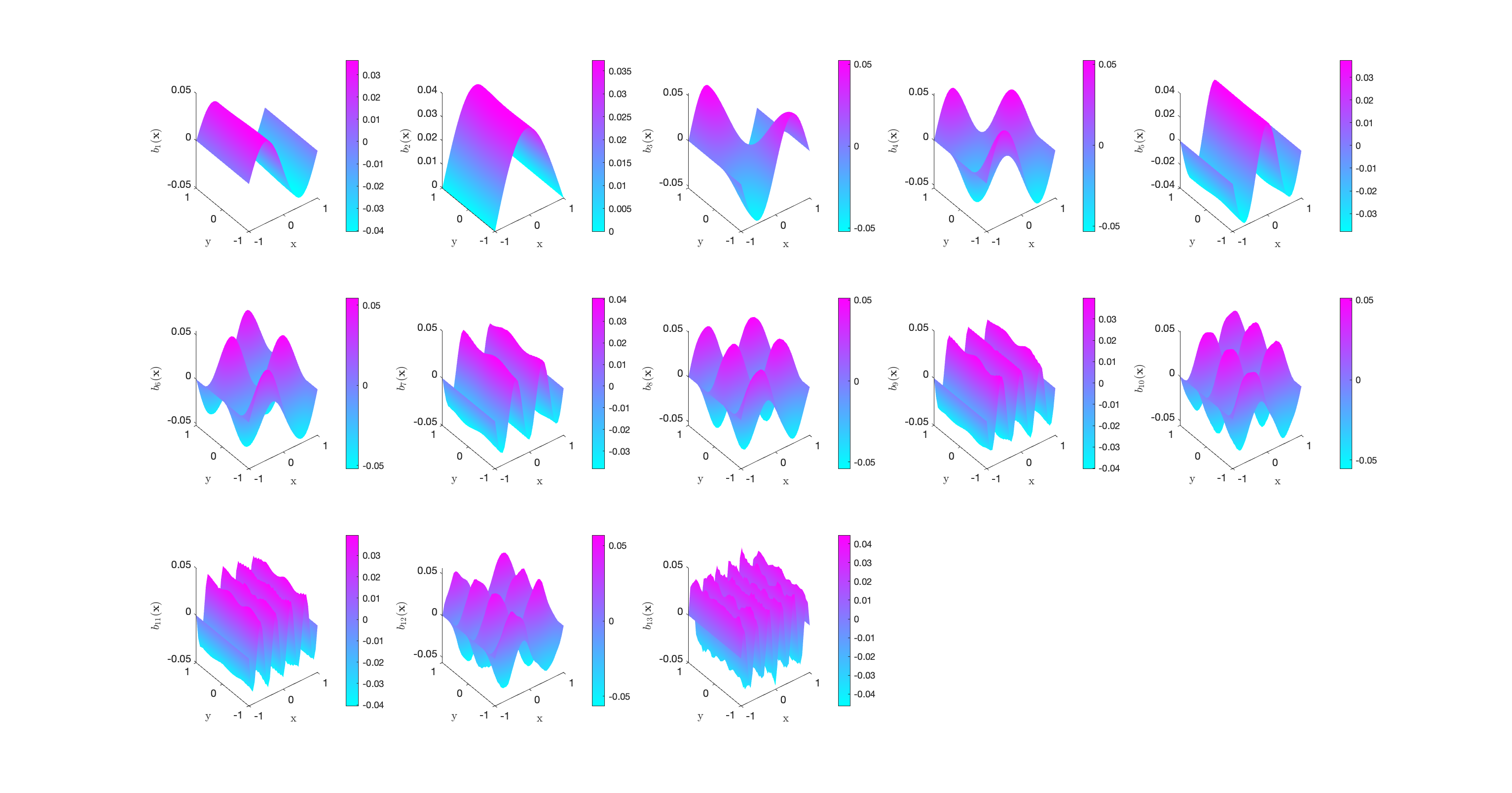}
		\caption{Ex. 5 (noiseless 2D wave): Reduced basis elements $b_j(\mathbf{x})$ for Fixed PC-FML.}
		\label{fig:ex3_noiseless_bases}
	\end{center}
\end{figure}

\begin{figure}[htbp]
	\begin{center}
        \includegraphics[width=.99\textwidth]{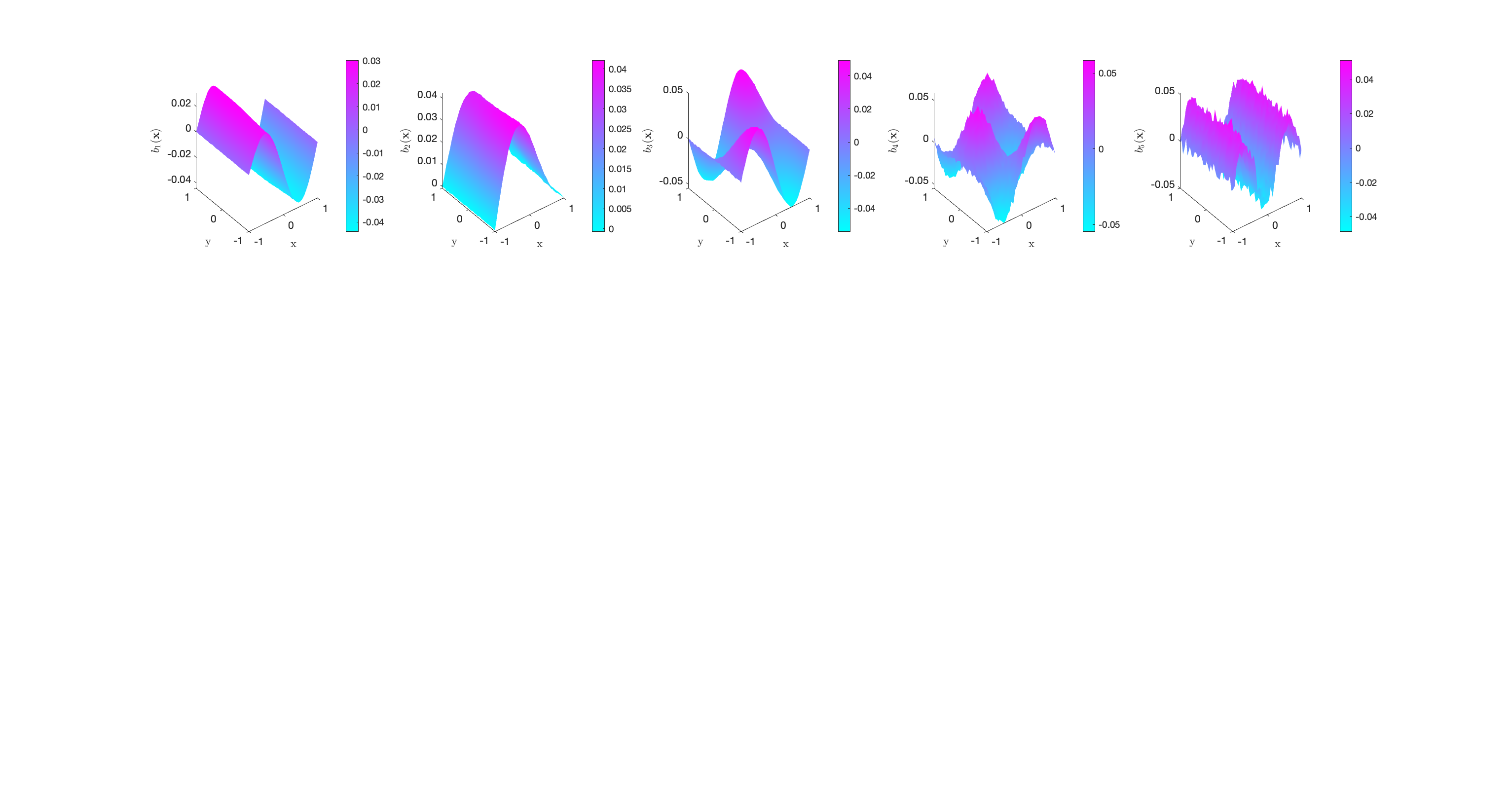}
		\caption{Ex. 5 (noisy 2D wave): Reduced basis elements $b_j(\mathbf{x})$ for Fixed PC-FML.}
		\label{fig:ex3_noisy_bases}
	\end{center}
\end{figure}

Figure \ref{fig:dimension} shows the decay of the singular values in each of the six forthcoming numerical experiments with and without noise. We see, e.g., in Ex. 5 the noiseless singular values of the training data drop off gradually after $N_{red}=13$, while noise is dominant after $N_{red}=5$ in the noisy observations. Additionally, this SVD computation exposes potential basis elements $b_j(\mathbf{x})$ for $j=1,\ldots,N_{red}$ in the columns of $V_{red}$, which can be used to explain the decomposition and interpret the dynamics. Figures \ref{fig:ex3_noiseless_bases} and \ref{fig:ex3_noisy_bases} show the 13- and 5-dimensional bases for the noiseless and noisy cases, respectively, in Example 3. We see similar shapes for the first 5 elements, while we also notice noise creeping into the $b_4(\mathbf{x})$ and $b_5(\mathbf{x})$ in the noisy measurements. This demonstrates why after this point, including extra basis elements is not advantageous as they will have a large noise component that may actually be detrimental to reconstructing the unknown dynamics.

\subsection{Model Training} \label{subsec:training}

With $N_{red}$ fixed, we consider three options for training the network, which correspond to options for choosing  $P_{in}\in\mathbb{R}^{N_{red}\times N_{full}}$ and $P_{out}\in\mathbb{R}^{N_{full}\times N_{red}}$. In all cases, the network operation is
\begin{align}
\begin{split}
\boldsymbol{\mathcal{N}}&\left(\mathbf{V}_n,\ldots,\mathbf{V}_{n-(N_{mem}-1)};\Theta,P_{in},P_{out}\right) =P_{out}P_{in}\mathbf{V}_n +
P_{out}\mathbf{M}\left(P_{in}\mathbf{V}_n,\ldots,P_{in}\mathbf{V}_{n-(N_{mem}-1)};\Theta\right)
\end{split}
\end{align}
To ease the notational burden, let the main component of the recurrent loss be
\begin{align}
L(\Theta,P_{in},P_{out}) = \frac1L\frac1K \sum_{l=1}^L\sum_{k=1}^K \|\boldsymbol{\mathcal{N}}^k(\mathbf{V}^{(l)}_n,\ldots,\mathbf{V}^{(l)}_{n-(N_{mem}-1)};\Theta)- \mathbf{V}^{(l)}_{n+k}\|_2^2,
\end{align}
where the power $k$ denotes composition of the network with itself $k$ times in the mode of \eqref{eq:prediction}. We consider the following options:

\begin{itemize}
\item \textbf{Fixed}: One of the main inspirations for this new method was to use a fixed linear transformation such as principal component analysis (PCA, see e.g. \cite{jolliffe2016principal}) to construct the maps $P_{in}$ and $P_{out}$. Therefore, consider $P_{in}$ and $P_{out}$ non-trainable and fixed in the following ``offline'' way. As in determining an appropriate reduced order dimension $N_{red}$ as in Section \ref{sec:chooserom}, compute the rank-$N_{red}$ truncated SVD of the data matrix $D$, $D_{red}=U_{red}\Sigma_{red}V_{red}^T$. Fix $P_{in}=V_{red}^T$ and $P_{out}=V_{red}$ and train the NN model \eqref{eq:model} with the dataset \eqref{eq:data_pdememory} by minimizimg the recurrent mean squared loss:
\begin{align}
\min_{\Theta} \left\{L(\Theta,V_{red}^T,V_{red})\right\}.
\end{align}
This fixed option allows us to control the modal basis.

\item \textbf{Constrained}: Consider $P_{in}$ and $P_{out}$ to be trainable, but constrained to be transposes of one another. Furthermore, impose regularization to encourage orthogonality, that $P_{in}P_{out}\approx I_{N_{red}\times N_{red}}$. In this case, we train the NN model \eqref{eq:model} with the dataset \eqref{eq:data_pdememory} by minimizing the recurrent mean squared loss balanced with a regularization penalty:
\begin{align}
\begin{split}
\min_{\Theta,P_{in},P_{out}} &\left\{L(\Theta,P_{in},P_{out})+\frac{\lambda}{2}\left\| P_{in}P_{in}^T-{I}_{N_R\times N_R}\right\|_2^2\right\}\\
&\text{subject to}\quad P_{out}=P_{in}^T.
\end{split}
\end{align}
In numerical examples below, we fix $\lambda=10^{-2}$. This constrained option lets the data choose the appropriate modal bases subject to constraints that mimic those obtained in the SVD approach.

\item \textbf{Unconstrained}: Consider $P_{in}$ and $P_{out}$ to be unconstrained matrices of trainable weights. In this case, we train the NN model \eqref{eq:model} with the dataset \eqref{eq:data_pdememory} by minimizing the recurrent mean squared loss:
\begin{align}
\min_{\Theta,P_{in},P_{out}} \left\{L(\Theta,P_{in},P_{out})\right\}.
\end{align}
This unconstrained option relies completely on the data to choose an appropriate modal basis.
\end{itemize}
Essentially, the options correspond to the amount of freedom we give the data and the network to learn the transformation. Indeed the inspiration for the general method was the Fixed regime, which can be relaxed to the Constrained regime's orthogonality and symmetry or completely dropped in the Unconstrained regime.\footnote{We admit that, despite theoretical guarantees of the SVD, the network may find a better FML basis! \NoRev{However, as stated in examples below, as $N_{red}$ increases, such as in nonlinear PDEs, the Constrained and Unconstrained approaches struggle to simultaneously learn appropriate reduced bases and system dynamics. Considering they are only trying to learn linear bases, this provides further support for the avoidance of learning highly-parameterized nonlinear bases, e.g. via autoencoder-decoders.}} Notably, in all cases the loss is measured in the nodal domain. This is significant, and differs from \cite{chen2023deep,WuXiu_modalPDE}, since the network parameters are trained to minimize loss in the nodal domain. If instead training is only executed in the modal domain, expanding back out could cause larger errors in the nodal domain, which is the focus. 

\textcolor{red}{}

\subsection{Model Prediction} \label{subsec:prediction}
Once trained, i.e. once $\Theta^*$, $P_{in}^*$, $P_{out}^*$ have been chosen, we obtain a predictive model over the grid set $X_{N_{grid}}$ for the underlying unknown PDE, such that given a new initial condition $(\mathbf{V}_{N_{mem}},\ldots,\mathbf{V}_1)^T$ over $X_{N_{grid}}$, we have
\begin{align}\label{eq:prediction}
\begin{cases}
(\bar{\mathbf{V}}_{N_{mem}},\ldots,\bar{\mathbf{V}}_1) = (\mathbf{V}_{N_{mem}},\ldots,\mathbf{V}_1),\\
\bar{\mathbf{V}}_{N_{mem}+k} = \boldsymbol{\mathcal{N}}(\bar{\mathbf{V}}_{N_{mem}+k},\ldots,\bar{\mathbf{V}}_k;\Theta^*,P_{in}^*,P_{out}^*),\quad k=1,2,\ldots
\end{cases}
\end{align}
where again the bar denotes the model's approximation.
\section{Computational Studies} \label{sec:examples}

In this section, we present several challenging numerical examples to demonstrate the PC-FML approach for learning PDEs from incomplete information. In each, the true governing PDEs are in fact known, but are used only to generate training and testing data. Their knowledge does not facilitate the PC-FML model construction. In other words, this is a purely data-driven approach. Full specification of the data generation as well as network structure parameters are provided in each example.

In an effort to avoid hand-picking results, all examples share the following properties and parameters:
\begin{itemize}
    \item Time-step: $\Delta t=10^{-2}$ apart from Ex. 4 which uses $\Delta t=8\times 10^{-3}$.
    \item Memory: fixed at $N_{mem}=20$ steps.
    \item Recurrent loss: $N_{rec}=10$ steps \RevA{for linear PDEs, and $N_{rec}=20$ for nonlinear.}
    \item Trajectories: limited to $N_{traj}=100$ training trajectories.
    \item Noise: additive zero-mean Gaussian noise is considered at two noise levels, with standard deviations $\sigma=0$ and $\sigma=10^{-1}$ or $10^{-2}$, referred to generically as noiseless and noisy.
    \item PC-FML network: networks $\mathbf{M}$ have $3$ hidden layers with up to $60$ nodes per layer determined by the multiple of $10$ above the reduced dimension. E.g., in Ex. 2 we have $N_{red}=14$ and so $20$ nodes are used in each layer. Hyperbolic tangent is used as the activation function.
    \item Nodal FML network size: comparison networks from \cite{churchill2023dnn} have $5$ disassembly channels with $1$ hidden layer with either $N_{grid}$ or $N_{red}$ nodes per layer\NoRev{, where using $N_{red}$ represents a naive attempt at dimensionality reduction.} The assembly layer has $5$ nodes. Hyperbolic tangent is used as the activation function.
    \item Optimization: in all cases, the Adam optimizer \cite{Adam_2014} is used to train the networks for $10,000$ epochs using a fixed  learning rate of $10^{-3}$.
    \item Ensemble prediction: both PC-FML and nodal FML use ensemble prediction \cite{churchill2022robust}, which sequentially uses an average of $10$ trained models as the input to predict the next time step.
\end{itemize}

\begin{table}
\centering
\begin{tabular}{ |c|c|c|c|c| } 
 \hline
Problem & Model & $N_{red}$ & $\sigma$ & $N_{params}$ \\ 
 \hline
\multirow{4}{*}{\makecell{Example 1\\(Heat)}} & Unconstrained PC-FML & $2$ &$0,10^{-1}$ &\hfill $1052$ \\ 
 & Constrained PC-FML & $2$ & $0,10^{-1}$ & \hfill  $852$\\
 & Fixed PC-FML& $2$ & $0,10^{-1}$ & \hfill $652$ \\
 & Nodal FML& N/A & $0,10^{-1}$ &\hfill $151036$ \\
 \hline
 \multirow{8}{*}{\makecell{Example 2\\ (Burgers')}} & Unconstrained PC-FML & $14$ & $0$ &\hfill $15154$ \\ 
 & Constrained PC-FML & $14$ & $0$ &\hfill $10954$ \\
 & Fixed PC-FML& $14$ & $0$ & \hfill $6754$ \\
 & Nodal FML& N/A & $0$ &\hfill $442606$ \\
 \cline{2-5}
 & Unconstrained PC-FML & $6$ & $10^{-1}$ &\hfill $5096$ \\ 
 & Constrained PC-FML & $6$ & $10^{-1}$ &\hfill $3296$ \\
 & Fixed PC-FML& $6$ & $10^{-1}$ & \hfill $1496$ \\
 & Nodal FML& N/A & $10^{-1}$ &\hfill $190566$ 
 \\
 \hline
 \multirow{4}{*}{\makecell{Example 3\\(Wave 1D)}} & Unconstrained PC-FML & $5$ & $0,10^{-1}$ &\hfill $2575$ \\ 
 & Constrained PC-FML & $5$ & $0,10^{-1}$ &\hfill $2325$ \\
 & Fixed PC-FML& $5$ & $0,10^{-1}$ & \hfill $2075$ \\
 & Nodal FML& N/A & $0,10^{-1}$ &\hfill $263036$ \\
 \hline
  \multirow{4}{*}{\makecell{Example 4\\(Shallow Water)}}
 & Fixed PC-FML& $52$ & $0$ & \hfill $93752$ \\
 & Nodal FML& N/A & $0$ &\hfill $1093296$ \\
 \cline{2-5}
 & Fixed PC-FML& $6$ & $10^{-2}$ & \hfill $9891$ \\
 & Nodal FML& N/A & $10^{-2}$ &\hfill $232091$ \\
 \hline
 \multirow{8}{*}{\makecell{Example 5\\(Wave 2D)}} & Unconstrained PC-FML & $13$ &$0$ &\hfill $44565$ \\ 
 & Constrained PC-FML & $13$ &$0$&\hfill $24584$ \\
 & Fixed PC-FML& $13$ & $0$&\hfill  $4603$ \\
 & Nodal FML& N/A & $0$ &\hfill $2105791$ \\
 \cline{2-5}
 & Unconstrained PC-FML & $5$ &$10^{-1}$ & \hfill $17445$ \\ 
 & Constrained PC-FML & $5$ &$10^{-1}$& \hfill $9760$ \\
 & Fixed PC-FML& $5$ & $10^{-1}$ &  \hfill $2075$ \\
 & Nodal FML& N/A & $10^{-1}$ &\hfill $814671$ \\
 \hline
  \multirow{4}{*}{\makecell{Example 6\\(Navier-Stokes)}}
 & Fixed PC-FML& $46$ & $0$&\hfill  $53496$ \\
 & Nodal FML& N/A & $0$ &\hfill $12087766$ \\
 \cline{2-5}
 & Fixed PC-FML& $26$ & $10^{-2}$ &  \hfill $18296$ \\
 & Nodal FML& N/A & $10^{-2}$ &\hfill $6837666$ \\
 \hline
\end{tabular}
\caption{Comparison of FML model parametrizations.}
\label{table:1}
\end{table}

%We first present the case of missing variables, that is, when data are available only for a subset of the state variables. The examples include a system of linear wave equations, as well as FitzHugh-Nagumo system with diffusion in both one and two spatial dimensions. In both cases only one state variable is observed, and the other one remains unknown.

%We then present the missing domain case, that is, when state variable data are available only in a sub-domain. Our examples include a fourth-order PDE, advection-diffusion equation, and viscous and inviscid Burger's equation. The parts of the domain where data are missing range from $10\%$ to as large as $60\%$.

\subsection{Example 1: 1D heat equation on incomplete non-uniform grid}
We first consider the heat equation
\begin{align}\label{eq:heat}
    \pi^2u_t &= u_{xx}
\end{align}
on the domain $\Omega=[0,1]$ with zero Dirichlet boundary conditions. The overall behavior of the PDE is characterized by dissipation.

To generate training data, first we solve the PDE numerically from $N_{traj}=100$ randomized initial conditions of the form
\begin{align}\label{eq:heatIC}
    u(x,0) = \alpha_1\sin(\pi x)+\alpha_2\sin(2\pi x)
\end{align}
where $\alpha_1,\alpha_2\sim U[-1,1]$, until $T=2$ ($200$ time steps). One chunk of length $N_{mem}+N_{rec}$ is selected randomly from each trajectory to form the training data set.  The solutions are observed on a randomly-chosen non-uniform grid of $N_{grid}=100$ points representing roughly the middle $50\%$ of the domain, $[.2399,.7577]\subset\Omega$.

The method from Section \ref{sec:chooserom} is used to determine an appropriate reduced basis of dimension $N_{red}=2$ for both $\sigma=0,10^{-1}$. Figure \ref{fig:dimension} shows the details. As indicated by the training data generation, $b_1(x) = \sin(\pi x)$ and $b_2(x) = \sin(2\pi x)$ represent optimal basis functions, however this fact is not used in the PC-FML approach.

\begin{figure}[htbp]
	\begin{center}
		\includegraphics[width=.99\textwidth]{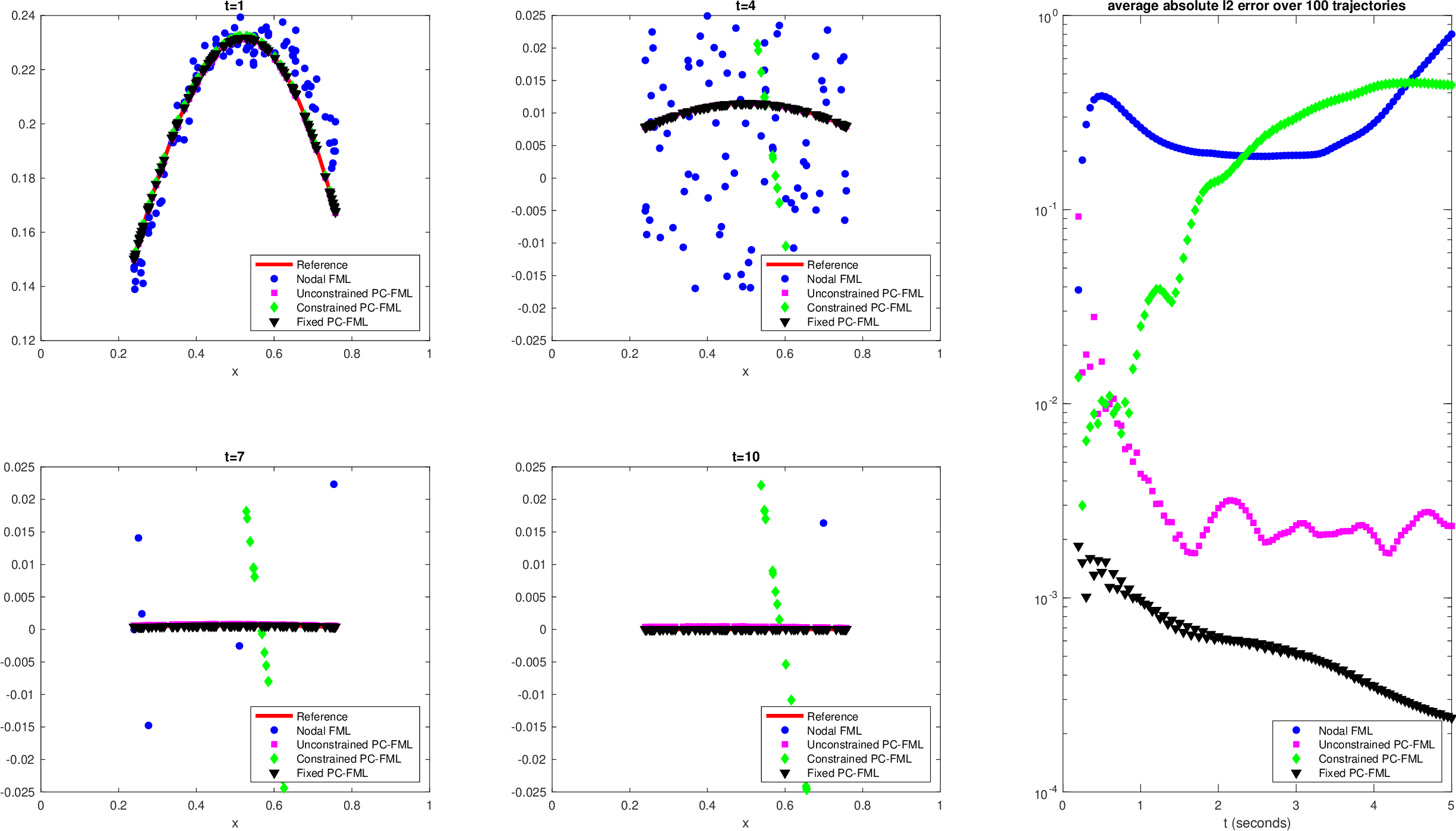}
		\caption{Ex. 1 (noiseless 1D heat): Example trajectory (left, center) and average error (right).}
		\label{fig:heat}
	\end{center}
\end{figure}

\begin{figure}[htbp]
	\begin{center}
		\includegraphics[width=.99\textwidth]{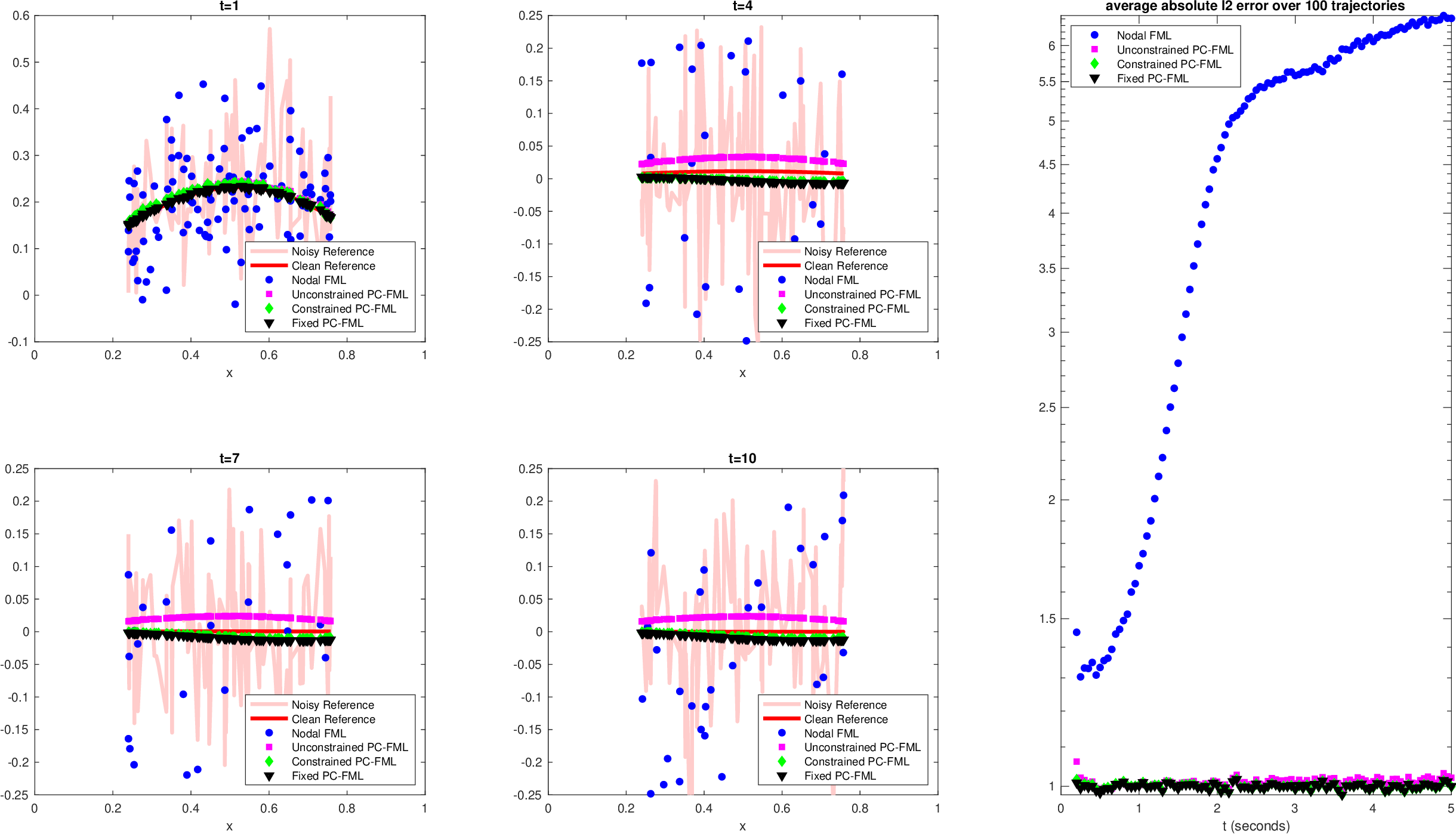}
		\caption{Ex. 1 (noisy 1D heat): Example trajectory (left, center) and average error (right).}
		\label{fig:heat1}
	\end{center}
\end{figure}

Validation testing also uses $100$ trajectories drawn similarly to \eqref{eq:heatIC}. On the left and center of Figures \ref{fig:heat} (noiseless) and \ref{fig:heat1} (noisy), an example test trajectory is shown at various prediction times, and on the right the average absolute $\ell_2$ error the test trajectories is shown over $500$ time steps. We see that the Fixed and Unconstrained PC-FML strategies stay stable and smooth over the entire prediction outlook, while the nodal FML deteriorates rapidly. Interestingly, we see the Constrained PC-FML approach successful in the noisy case, while it appears to get ``stuck'' after a short time in the noiseless experiment. We hypothesize that perhaps without a good initialization (no special setting is used here), the Constrained approach fails to find an appropriate basis that abides by its constraints. Additionally, the denoising behavior of PC-FML approach is evident.

We note the success of these methods out until $T=10$ despite having only ``seen'' dynamics up to $T=2$ in the training data. Table \ref{table:1} shows that the failing nodal FML model uses $151036$ parameters, while the more successful PC-FML models use just $1052$, $852$, and $652$ for the Unconstrained, Constrained, and Fixed options, respectively.

\subsection{\RevA{Example 2: 1D Burgers' equation on incomplete uniform grid}}

We next consider the nonlinear Burgers' equation with viscosity
\begin{align}\label{eq:burgers}
    u_t + uu_x &= 0.05 u_{xx}
\end{align}
on the domain $\Omega=[-\pi,\pi]$ with periodic boundary conditions. The overall behavior of the PDE is characterized by formation of a shock but then dissipation before a discontinuity forms.

To generate training data, first we numerically solve the PDE from $N_{traj}=100$ randomized initial conditions as in \eqref{eq:heatIC} where $\alpha_1,\alpha_2\sim U[-1,1]$, until $T=3$ ($300$ observed time steps). One chunk of length $N_{mem}+N_{rec}$ is selected randomly from each trajectory to form the training data set. The solutions are observed on a uniform grid of $N_{grid}=300$ points representing roughly the middle $60\%$ of the domain, $[-3\pi/5,3\pi/5]\subset\Omega$.

The method from Section \ref{sec:chooserom} is used to determine an appropriate reduced basis of dimension $N_{red}=14$ for $\sigma=0$ and $N_{red}=6$ for $10^{-1}$. Figure \ref{fig:dimension} shows the details.

\begin{figure}[htbp]
	\begin{center}
    \includegraphics[width=.99\textwidth]{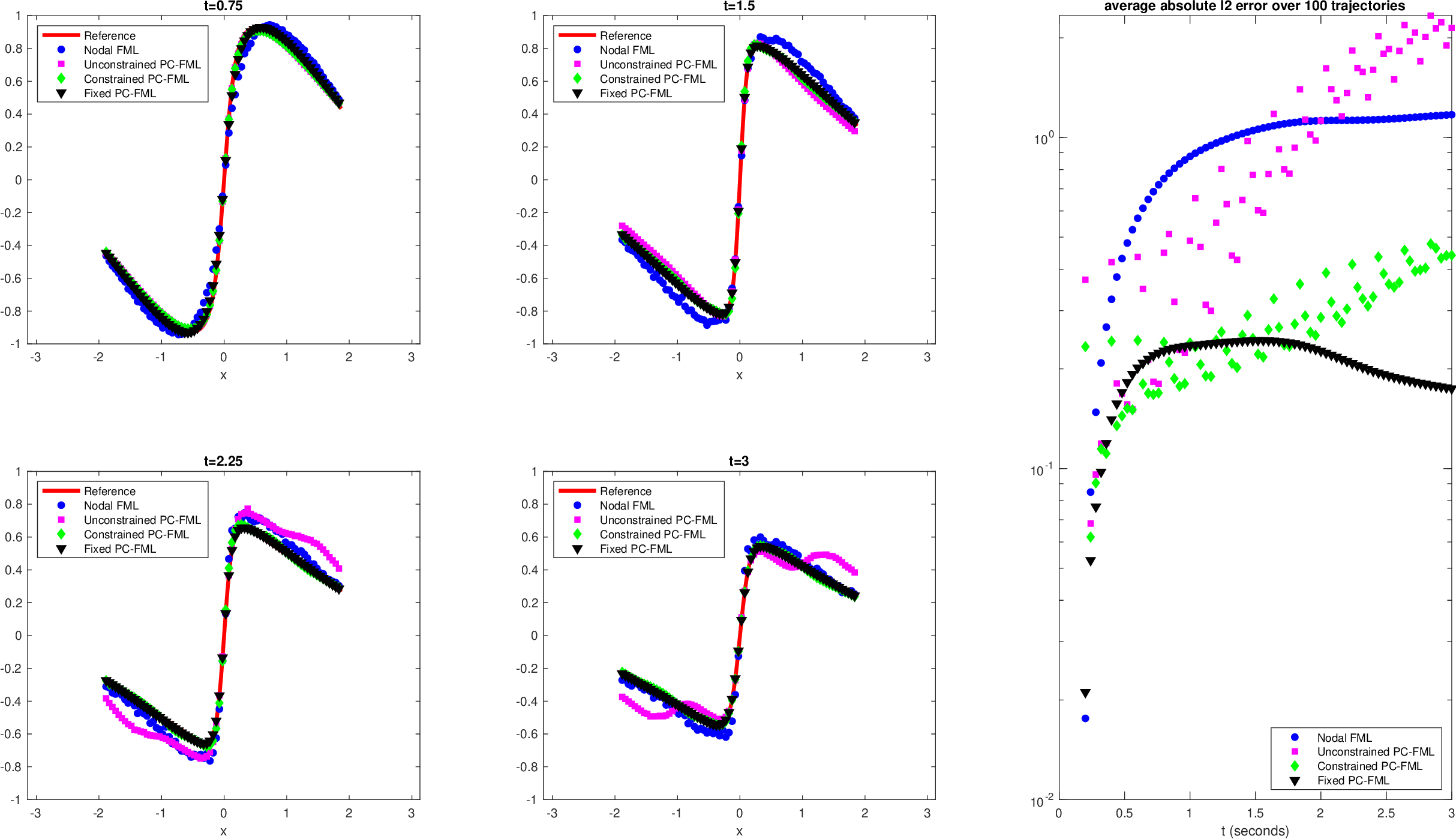}
		\caption{Ex. 2 (noiseless 1D Burgers'): Example trajectory (left, center) and average error (right).}
		\label{fig:burgers}
	\end{center}
\end{figure}

\begin{figure}[htbp]
	\begin{center}
	\includegraphics[width=\textwidth]{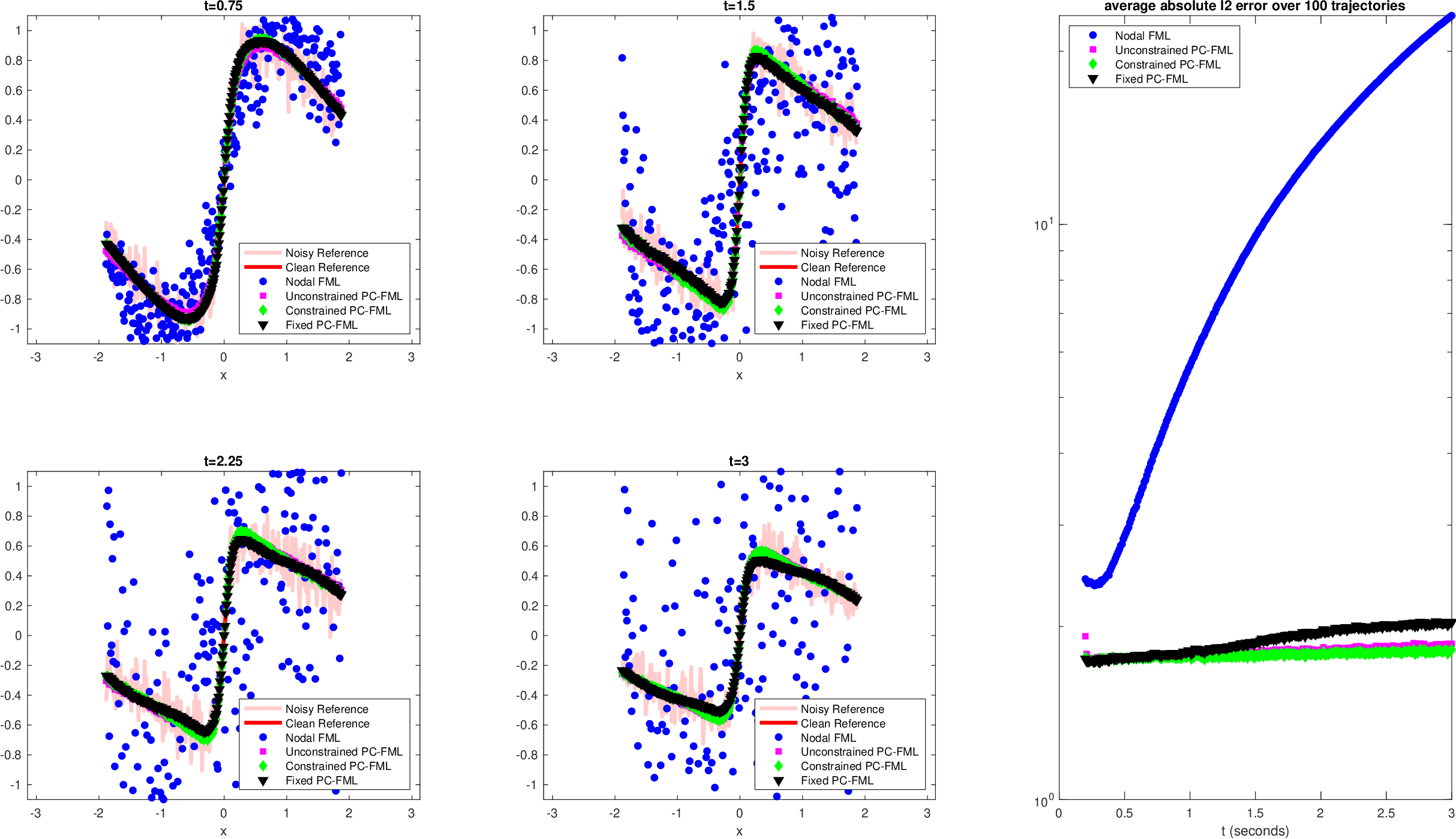}
		\caption{Ex. 2 (noisy 1D Burgers'): Example trajectory (left, center) and average error (right).}
		\label{fig:burgers_noisy}
	\end{center}
\end{figure}

Validation testing also uses $100$ trajectories drawn similarly to \eqref{eq:heatIC}. On the left and center of Figures \ref{fig:burgers} (noiseless) and \ref{fig:burgers_noisy} (noisy), an example test trajectory is shown at various prediction times, and on the right the average absolute $\ell_2$ error the test trajectories is shown over $300$ time steps. In the noiseless case, the Fixed PC-FML strategy performs best. We hypothesize that because the $N_{red}=14$ is relatively high (representing a $N_{red}N_{mem}=280$ element input to $\mathbf{M}$), that Constrained the Unconstrained PC-FML have the additional challenge of discovering a basis rather than having one provided. Interestingly, all the PC-FML strategies are successful in the noisy case at learning denoised evolution of the system. Table \ref{table:1} shows that the nodal FML models use between 29 and 127 times more parameters compared with PC-FML.

\subsection{Example 3: 1D wave system with missing variables} \label{sec:wave}

Next we consider the system
\be
\begin{bmatrix} v \\ w \end{bmatrix}_t = \begin{bmatrix} 0 & 1 \\ 1 & 0 \end{bmatrix} \begin{bmatrix} v \\ w \end{bmatrix}_x,
\ee
on the domain $\Omega = [-1,1]$ with periodic boundary conditions. We consider the case when
only data on the solution component $v$ are available (i.e. $N_{obs}=1$) and aim at
constructing an accurate NN predictive model for the evolution of $v$.

To generate training data, first we solve the PDE analytically until $T=1$ ($100$ time steps) from $N_{traj}=100$ randomized initial conditions given by
\begin{align}\label{eq:waveICs}
\begin{bmatrix} v \\ w \end{bmatrix}(x,0) = \begin{bmatrix} a_0+\sum_{n=1}^{N_b} (a_n\cos(nx)+b_n\sin(nx)) \\ c_0+\sum_{n=1}^{N_b} (c_n\cos(nx)+d_n\sin(nx))\end{bmatrix},
\end{align}
with $a_0,c_0\sim U[-\frac12, \frac12]$ and $a_n,b_n,c_n,d_n\sim
U[-\frac1n, \frac1n]$, and $N_b=2$. Then, one random chunk of length $N_{mem}+N_{rec}$ per trajectory from $v$ only are recorded into our training data. The solutions are observed on a uniform grid of $N_{grid}=50$ points.

The method from Section \ref{sec:chooserom} is used to determine an appropriate reduced basis of dimension $N_{red}=5$ for both the noiseless and noisy cases. Figure \ref{fig:dimension} (center) shows the details. As in Ex. 1, indicated by the training data generation, there are an obvious set of $5$ optimal basis functions, however this fact is not used.

\begin{figure}[htbp]
	\begin{center}
		\includegraphics[width=\textwidth]{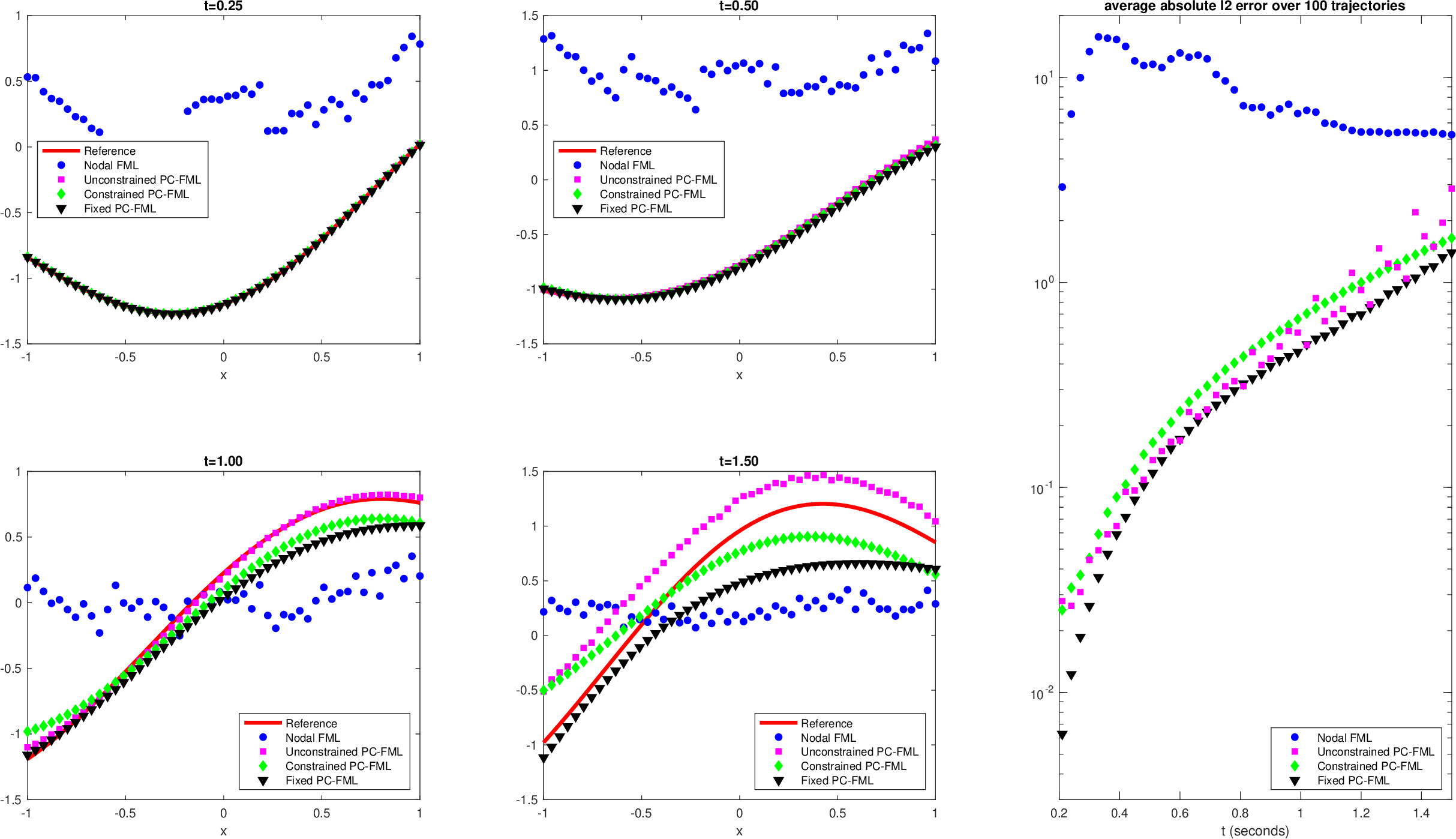}
		\caption{Ex. 3 (noiseless 1D wave): Example trajectory (left, center) and average error (right).}
		\label{fig:ex2_noiseless}
	\end{center}
\end{figure}

\begin{figure}[htbp]
	\begin{center}
		\includegraphics[width=\textwidth]{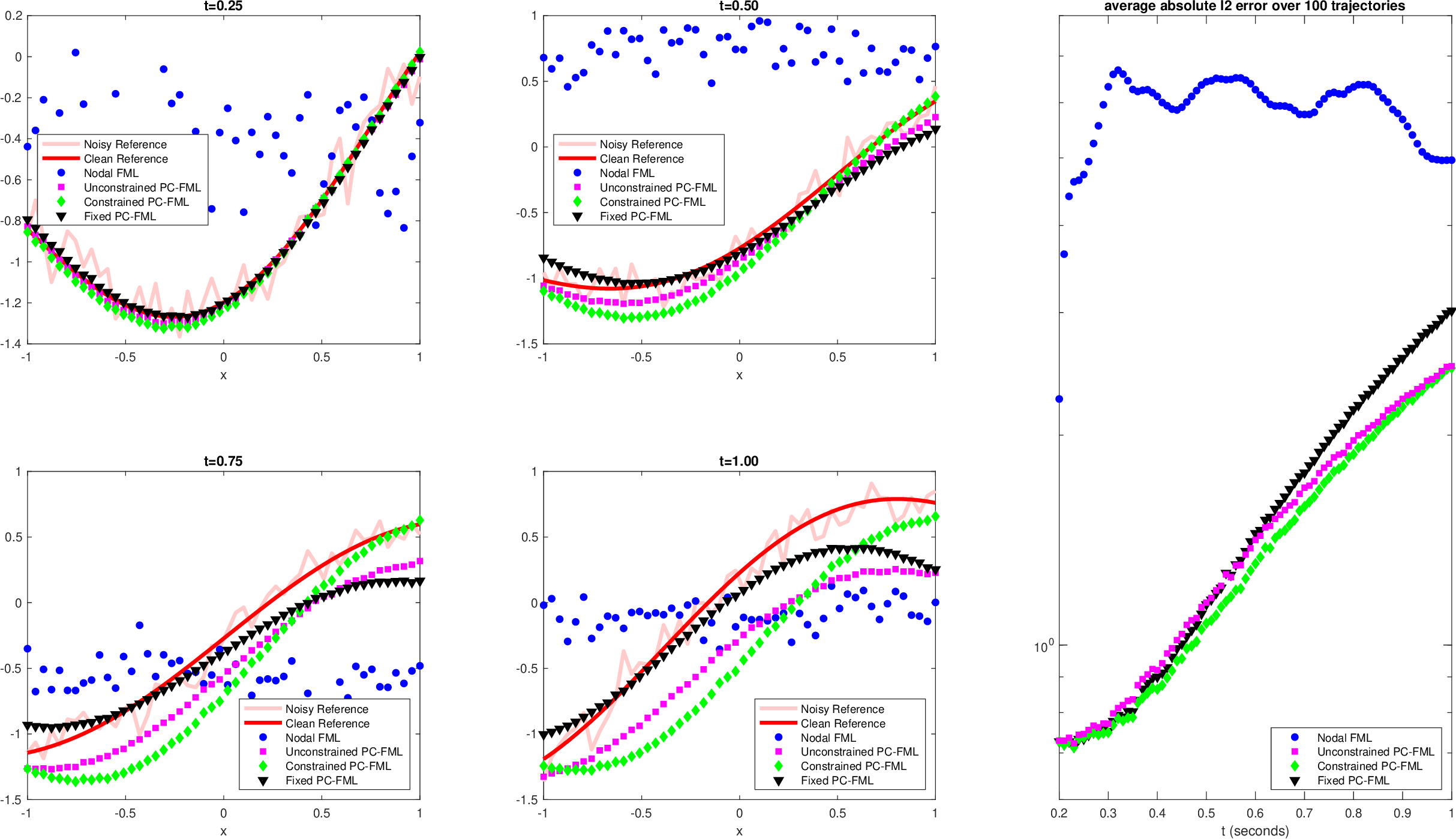}
		\caption{Ex. 3 (noisy 1D wave): Example trajectory (left, center) and average error (right).}
		\label{fig:ex2_noisy}
	\end{center}
\end{figure}

Validation testing also uses $100$ trajectories drawn similarly to \eqref{eq:waveICs}. On the left and center of Figures \ref{fig:ex2_noiseless} (noiseless) and \ref{fig:ex2_noisy} (noisy), an example test trajectory is shown at various prediction times, and on the right the average absolute $\ell_2$ error the test trajectories is shown over $300$ time steps. We see that the PC-FML strategies stay relatively stable and smooth over the prediction outlook, while the nodal FML deteriorates almost immediately. Particularly in the noisy case, deviations here are similar to those observed in numerical methods, whereby small errors in initial conditions accumulate and while the solution exhibits ``wave-like'' behavior, it is ``displaced'' from the appropriate trajectory. Additionally, the denoising behavior of PC-FML approach is evident.

Table \ref{table:1} shows that the nodal FML model uses $263036$ parameters, while the PC-FML models use just $2575$, $2325$, and $2075$ for the Unconstrained, Constrained, and Fixed options, respectively.

\subsection{\RevA{Example 4: 1D shallow water system with missing variables}}

Next we consider the nonlinear shallow water equations
\begin{align}
\begin{split}
h_t &+ (hu)_x = 0 \\
(hu)_t &+ \left(hu^2+\frac{9.8}{2}h^2\right)_x = 0
\end{split}
\end{align}
where $h$ and $u$ represent height and velocity of a fluid, respectively. We operate on the domain $\Omega = [0,10]$ with periodic boundary conditions. We consider the case when only data on the solution component $h$ are available (i.e. $N_{obs}=1$) and aim at constructing an accurate NN predictive model for the evolution of $h$.

To generate training data, first we numerically solve the PDE until $T=5$ ($626$ time steps) from $N_{traj}=100$ randomized initial conditions, representing a sinusoidal perturbation in the fluid height with zero initial velocity, given by
\begin{align}\label{eq:sweICs}
\begin{split}
h(x,0) &= 1+\alpha \sin\left(\frac{2\pi x}{10}\right), \quad u(x,0) =0
\end{split}
\end{align}
with $\alpha\sim U[0.1,1.0]$. Then, one random chunk of length $N_{mem}+N_{rec}$ per trajectory from $h$ only are recorded into our training data. The solutions are observed on a uniform grid of $N_{grid}=200$ points.

The method from Section \ref{sec:chooserom} is used to determine an appropriate reduced basis of dimension $N_{red}=52$ for the noiseless case and $N_{red}=6$ noisy case of $\sigma=10^{-2}$. Figure \ref{fig:dimension} shows the details. Particularly in this example, the dimension $N_{red}$ required to approximate solution data with high accuracy is quite large, resulting in $P_{in}$ and $P_{out}$ operators of size $200\times52$ and $52\times 200$ and presenting an input to the flow map network $\mathbf{M}$ of length $1040$. Because of these dimensionality challenges and our focus on rapid and inexpensive local simulation of scientific data, in this case we consider only the Fixed approach, thereby reducing the training burden on PC-FML.

\begin{figure}[htbp]
	\begin{center}
	\includegraphics[width=\textwidth]{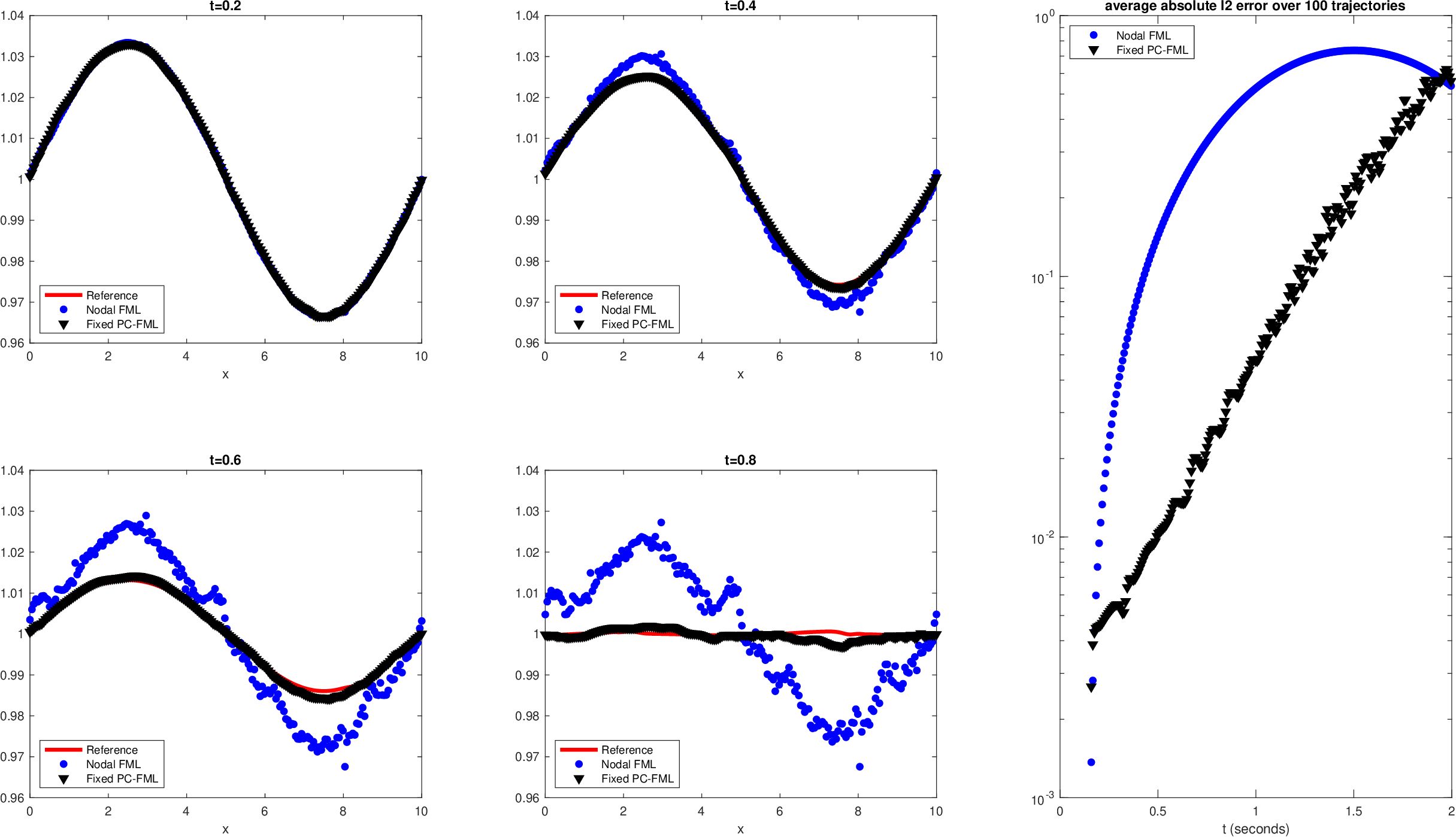}
		\caption{Ex. 4 (noiseless 1D shallow water): Example trajectory (left, center) and average error (right).}
		\label{fig:swe_noiseless}
	\end{center}
\end{figure}

\begin{figure}[htbp]
	\begin{center}
	\includegraphics[width=\textwidth]{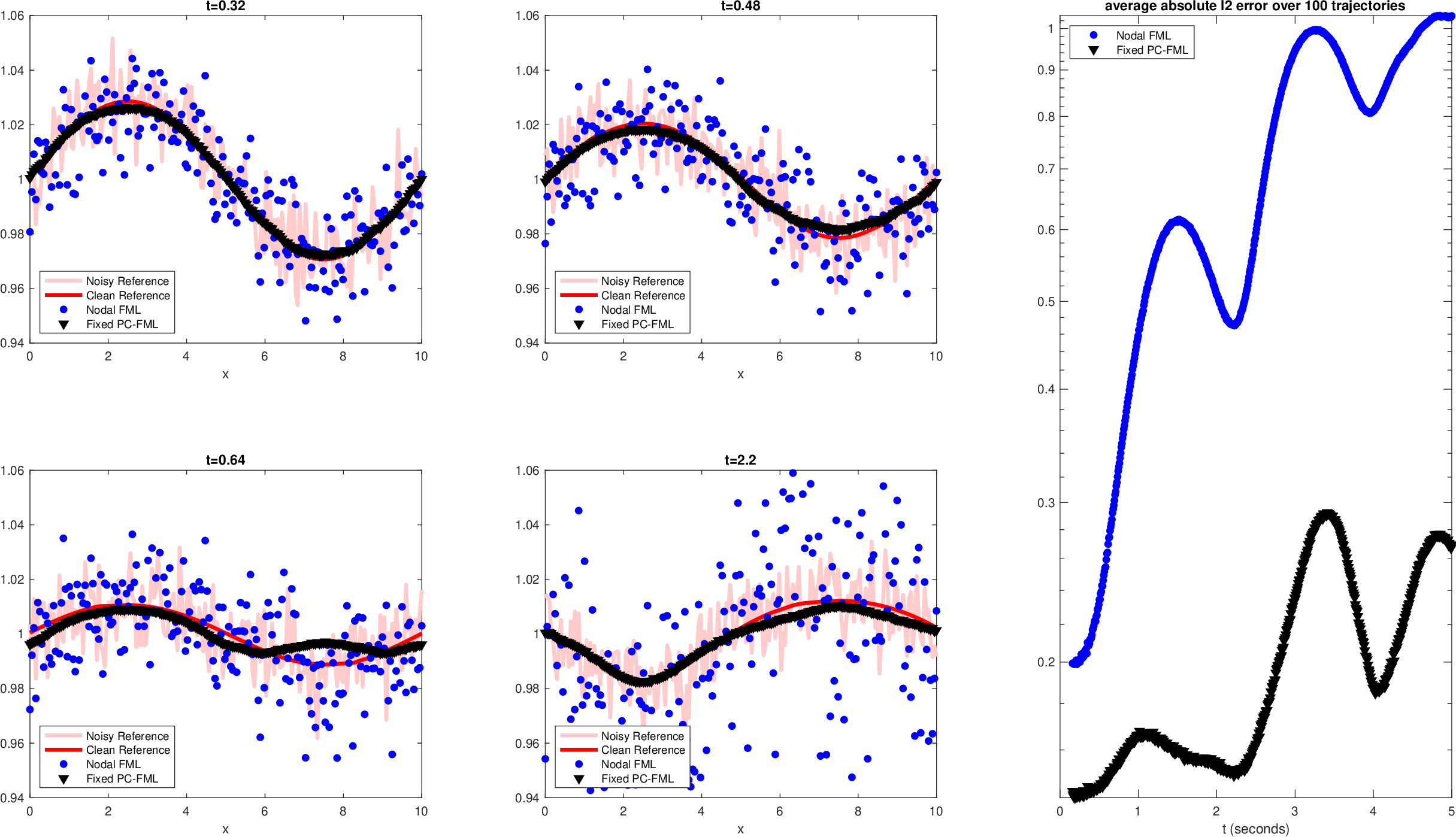}
		\caption{Ex. 4 (noisy 1D shallow water): Example trajectory (left, center) and average error (right).}
		\label{fig:swe_noisy}
	\end{center}
\end{figure}

Validation testing also uses $100$ trajectories drawn similarly to \eqref{eq:sweICs}. On the left and center of Figures \ref{fig:swe_noiseless} (noiseless) and \ref{fig:swe_noisy} (noisy), an example test trajectory is shown at various prediction times, and on the right the average absolute $\ell_2$ error the test trajectories is shown over $100$ and $275$ time steps, respectively. In the noiseless case, we see that the nodal FML model fails to evolve, while the Fixed PC-FML model follows the correct evolution before being infected with small perturbations in the prediction. In the noisy case, the Fixed PC-FML performs even better, with denoised predictions with some accuracy up until $t=2.2$. Since the shallow water equations represent a hyperbolic conservation law, we expect and plan that improvements to these results can be made by integrating into PC-FML techniques from \cite{chen2024learning,liu2024entropy} that enforce conservation.

Table \ref{table:1} shows that the nodal FML model uses between $10$ and $25$ times more parameters than the PC-FML approach.

\subsection{Example 5: 2D wave equation on non-uniform grid with missing information} \label{sec:wave2D}

\NoRev{In our first two-dimensional example}, we consider the wave equation
\begin{align}
    u_{tt} = \Delta u
\end{align}
on the domain $\Omega = [-1,1]^2$ with zero Dirichlet vertical boundary conditions (i.e. $x=-1,1$) and zero Neumann horizontal boundary conditions (i.e. $y=-1,1$). We note that in this form the PDE in fact does not conform to \eqref{govern} since it is second order in time. This is therefore indeed a problem of a ``missing variable'' (i.e. $u_t$), such that observing $u$ alone is partial information for the model.

\begin{figure}[htbp]
	\begin{center}
		\includegraphics[width=.5\textwidth]{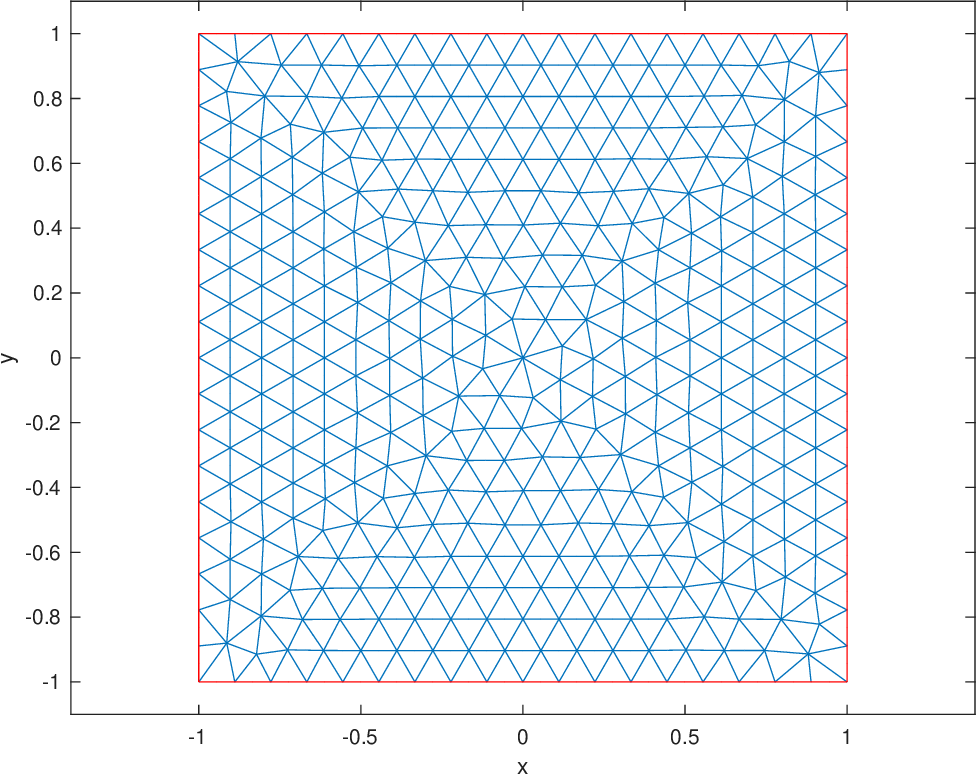}
		\caption{Ex. 5 (2D wave): Non-uniform measurement grid.}
		\label{fig:ex3_grid}
	\end{center}
\end{figure}

To generate training data, first we numerically solve the PDE until $T=4$ ($400$ time steps) from $N_{traj}=100$ randomized initial conditions given by
\begin{align}\label{eq:wave2ICs}
\begin{split}
u(x,0) &= \arctan(\alpha_1\cos((\pi/2)x)) \\
u_t(x,0) &= \alpha_2\sin(\pi x)\exp(\alpha_3\sin((\pi/2) y))
\end{split}
\end{align}
with $\alpha_1,\alpha_3\sim U[-1, 1]$ and $\alpha_2\sim U[-3,3]$. Then, one random chunk of length $N_{mem}+N_{rec}$ per trajectory from $u$ are recorded into the training data. The solutions are observed on a non-uniform grid of $N_{grid}=1537$ points shown in Figure \ref{fig:ex3_grid}.

The method from Section \ref{sec:chooserom} is used to determine an appropriate reduced basis of dimension $N_{red}=13$ for the noiseless case $(\sigma=0)$ and $N_{red}=5$ for the noisy case $(\sigma=10^{-1})$. Figure \ref{fig:dimension} (right) shows the details. In particular, we see that the noiseless singular values of the training data drop off gradually after $N_{red}=13$, while noise is dominant after $N_{red}=5$ in the noisy observations. Unlike Exs. 1 and 2, indicated by the training data generation, there are not obvious optimal basis functions for the initial conditions. The Fixed PC-FML approach provides insight, whereby Figures \ref{fig:ex3_noiseless_bases} and \ref{fig:ex3_noisy_bases} expose the 13- and 5-dimensional bases $b_j(\mathbf{x})$ for $j=1,\ldots,N_{red}$ in the columns of $V_{red}$ for the noiseless and noisy cases, respectively. We see similarities in the first basis 5 elements, but also notice noise creeping into the $b_4(\mathbf{x})$ and $b_5(\mathbf{x})$ in the noisy measurements. This demonstrates why after this point, including extra basis elements is not advantageous as they will have a large noise component that may actually be detrimental to reconstructing and explaining the unknown dynamics. The bases discovered by the Unconstrained and Constrained can be analyzed in the same way.

\begin{figure}[htbp]
	\begin{center}
		\includegraphics[width=\textwidth]{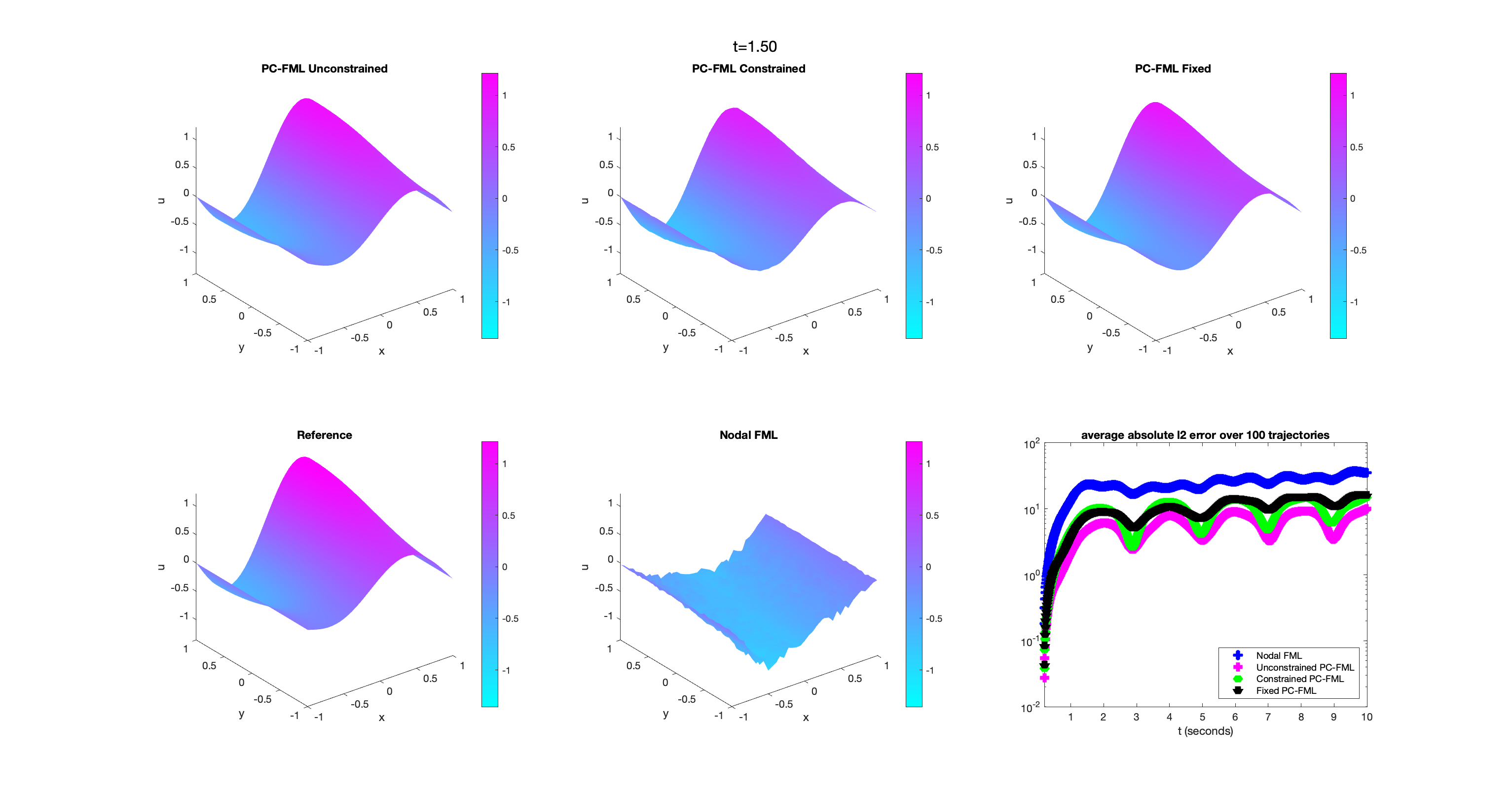}
		\caption{Ex. 5 (noiseless 2D wave): Example trajectory (left, center) and average error (right).}
		\label{fig:ex3_noiseless}
	\end{center}
\end{figure}

\begin{figure}[htbp]
	\begin{center}
		\includegraphics[width=\textwidth]{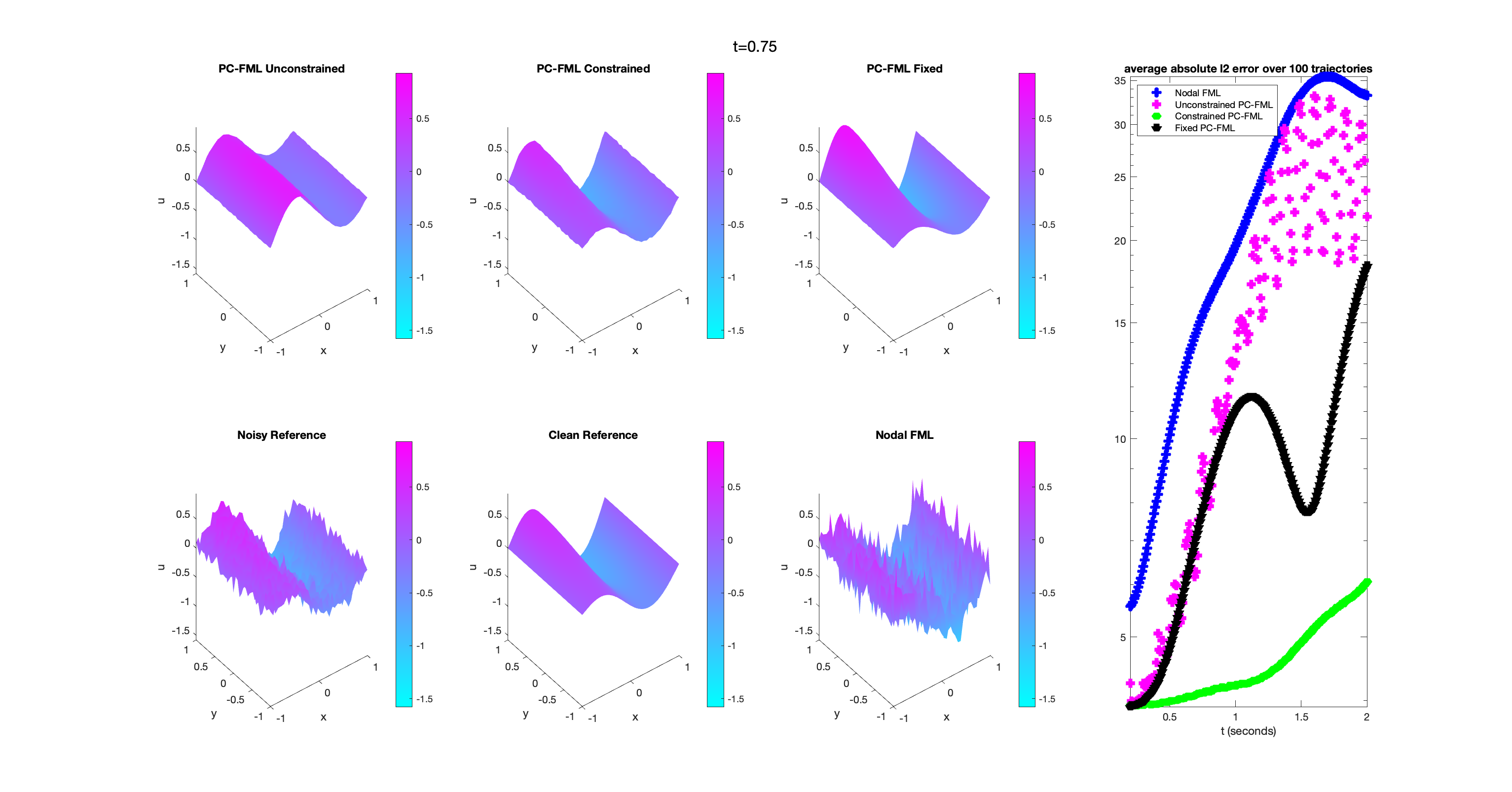}
		\caption{Ex. 5 (noisy 2D wave): Example trajectory (left, center) and average error (right).}
		\label{fig:ex3_noisy}
	\end{center}
\end{figure}

Validation testing also uses $100$ trajectories drawn similarly to \eqref{eq:wave2ICs}. On the left and center of Figures \ref{fig:ex3_noiseless} (noiseless) and \ref{fig:ex3_noisy} (noisy), an example test trajectory is shown at various prediction times, and on the right the average absolute $\ell_2$ error the test trajectories is shown over $1000$ and $200$ time steps, respectively. We see that the PC-FML strategies stay relatively stable and smooth over the prediction outlook, while the nodal FML noisily deteriorates almost immediately and never actually exhibits ``wave-like'' behavior. Additionally, the denoising behavior of PC-FML approach is evident.

Table \ref{table:1} shows for the noiseless case, the nodal FML model uses $2105791$ parameters, while the PC-FML models use just $44565$, $24584$, and $4603$ for the Unconstrained, Constrained, and Fixed options, respectively. In the noisy case, the nodal FML model uses $814671$ parameters, while the PC-FML models use just $17445$, $9760$, and $2075$ for the Unconstrained, Constrained, and Fixed options, respectively.

\subsection{\RevA{Example 6: 2D Navier-Stokes equation}}

Finally, we consider the nonlinear Navier-Stokes equations in their vorticity-streamfunction formulation
\begin{align}
\begin{split}
    \omega_t + \psi_y \omega_x-\psi_x\omega_y &= \nabla^2\omega \\ 
    \nabla^2\psi +\omega &= 0
\end{split}
\end{align}
where $\omega$ represents vorticity and $\psi$ the streamfunction. We operate on the domain $\Omega = [0,10]^2$ with periodic boundary conditions in both directions. Since $\psi$ is directly accessible by solving a Laplace equation if $\omega$ is known, observing only $\omega$ represents a fully-observed system. We aim at constructing an accurate NN predictive model for the evolution of $\omega$, demonstrating the value of PC-FML in the fully-observed case as well.

To generate training data, first we numerically solve the PDE until $T=5$ ($500$ time steps) from $N_{traj}=100$ randomized initial conditions given by
\begin{align}\label{eq:nsICs}
\begin{split}
\omega(x,0) &= e^{-\frac12\left((x - x_0)^2 + (y - y_0)^2\right)}
\end{split}
\end{align}
with $x_0,y_0\sim U[2,8]^2$. Then, one random chunk of length $N_{mem}+N_{rec}$ per trajectory from $u$ are recorded into the training data. The solutions are observed on a uniform grid of $N_{grid}=50^2$ points.

The method from Section \ref{sec:chooserom} is used to determine an appropriate reduced basis of dimension $N_{red}=46$ for the noiseless case $(\sigma=0)$ and $N_{red}=26$ for the noisy case $(\sigma=10^{-2})$. Figure \ref{fig:dimension} shows the details. In particular, we see that the noiseless singular values of the training data drop off gradually after $N_{red}=46$, while noise is dominant after $N_{red}=26$ in the noisy observations. Unlike the linear examples, indicated by the training data generation, there are not obvious optimal basis functions for the initial conditions. Like in Ex. 4, we focus on the Fixed PC-FML approach due to the high dimensionality even in the reduced basis.

\begin{figure}[h]
	\begin{center}
		\includegraphics[width=\textwidth]{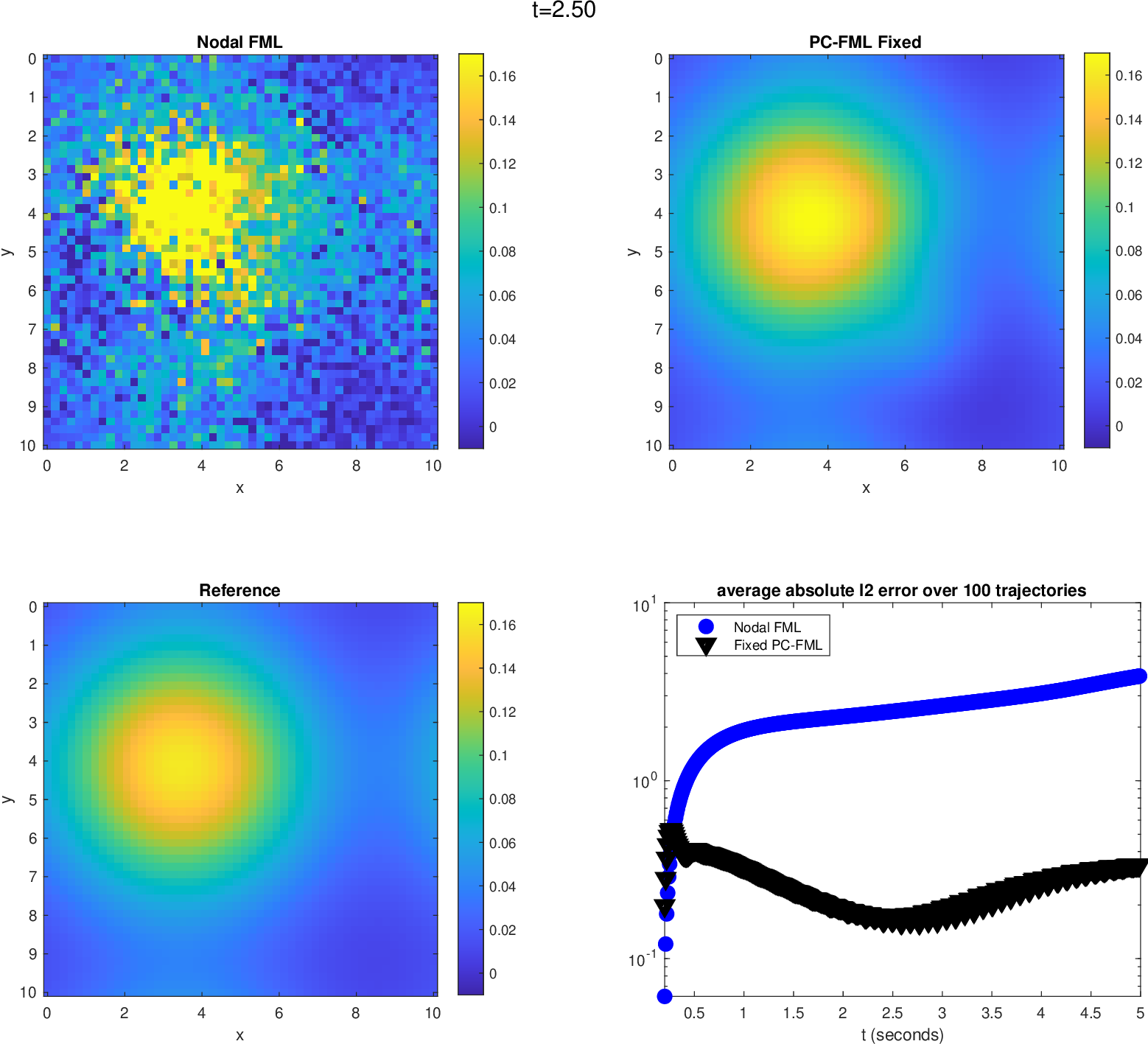}
		\caption{Ex. 6 (noiseless Navier-Stokes): Example vorticity fields at $t=2.5$ and average error (bottom right).}
		\label{fig:ns_noiseless}
	\end{center}
\end{figure}

\begin{figure}[h]
	\begin{center}
		\includegraphics[width=\textwidth]{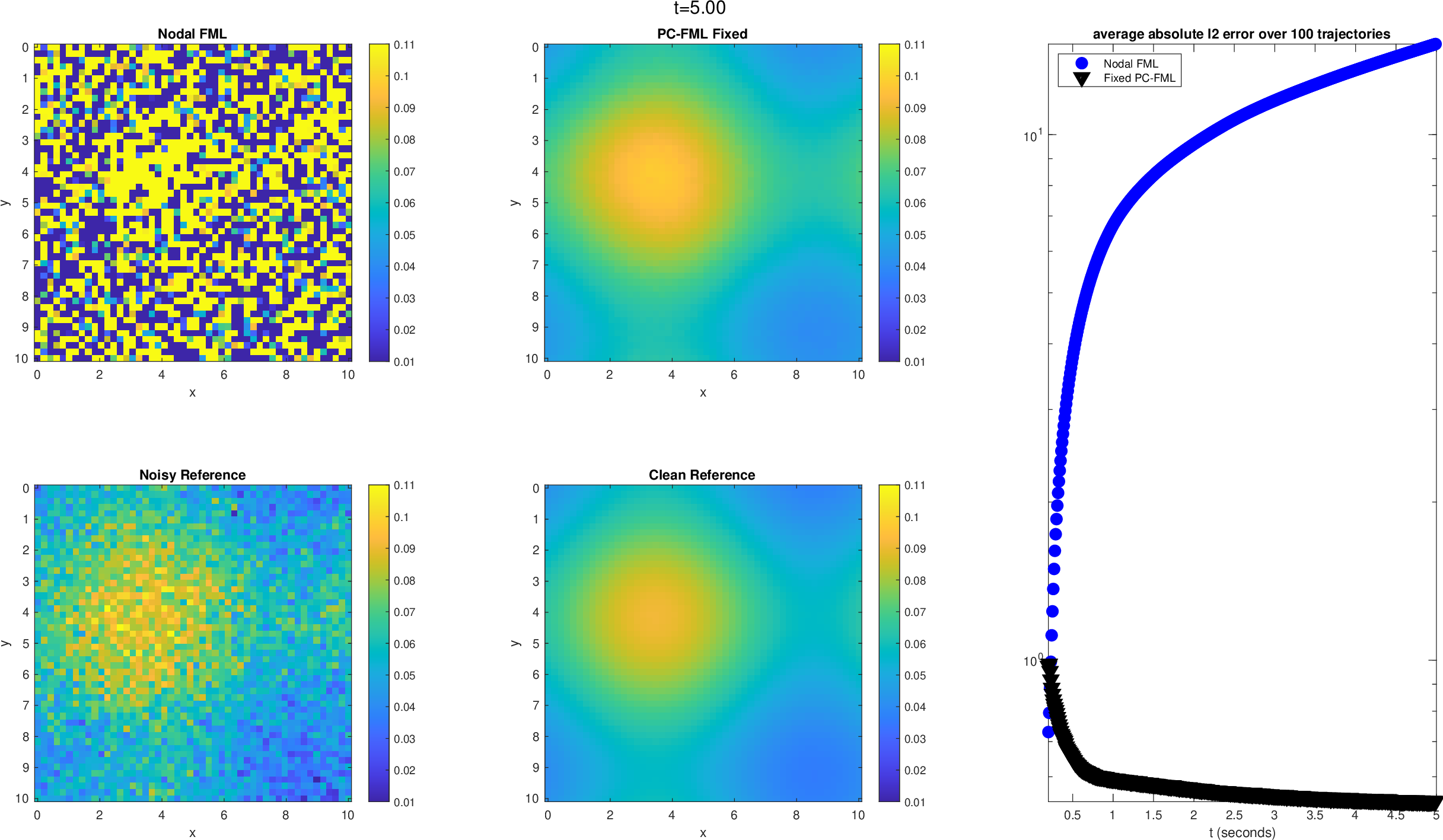}
		\caption{Ex. 6 (noisy Navier-Stokes): Example vorticity fields at $t=5$ (center, left) and average error (right).}
		\label{fig:ns_noisy}
	\end{center}
\end{figure}

Validation testing also uses $100$ trajectories drawn similarly to \eqref{eq:nsICs}. On the left and center of Figures \ref{fig:ns_noiseless} (noiseless) and \ref{fig:ns_noisy} (noisy), an example test trajectory is shown at various prediction times, and on the (bottom) right the average absolute $\ell_2$ error the test trajectories is shown over $500$ time steps. Fixed PC-FML stays stable and smooth over the prediction outlook, while the nodal FML noisily deteriorates almost immediately, with the comparison particularly stark in the noisy case. Generally, the results show that despite its simplicity, the SVD-motivated PC-FML approach yields accurate approximation to data resulting from a nonlinear PDE.

Table \ref{table:1} shows for the noiseless case, the nodal FML model uses $12087766$ parameters, while the Fixed PC-FML models use just $53496$. In the noisy case, the nodal FML model uses $6837666$ parameters, while the Fixed PC-FML models use just $18296$, representing a compression of over $300$ times.

\section{Conclusion} \label{sec:conclusions}

We have presented a computational technique in the Flow Map Learning (FML) family for modeling the evolution of unknown PDEs from limited, noisy, and incomplete solution data using a novel NN architecture with significantly reduced parameterization, notably enabling hundredfold lower training data requirements \RevA{and the potential for rapid local simulation of high-dimensional systems}. In particular, the architecture achieves this drastic decrease in parameters by first learning a common reduced linear approximation to reduce high-dimensional spatial measurements to a few important components, and then conducting the dynamics learning in this reduced basis which requires a lower complexity network due to its ODE form. Future work will focus on \NoRev{incorporating physics-based constraints to further reduce representations and dynamics} in the context of data-driven FML of PDEs.
\bibliographystyle{siamplain}
\bibliography{neural,LearningEqs,ensemble}

\end{document}